\documentclass[runningheads]{llncs}

\usepackage[FINAL,year=2024,ID=2468]{eccv} %

\usepackage{nameref,zref-xr}
\zxrsetup{toltxlabel}
\zexternaldocument*{supplementary}

\usepackage{eccvabbrv}

\usepackage{graphicx}
\usepackage{booktabs}
\usepackage{multirow}
\usepackage{makecell}
\usepackage{algorithm}
\usepackage{algorithmic}
\usepackage[accsupp]{axessibility}  

\usepackage[pagebackref,breaklinks,colorlinks,citecolor=eccvblue]{hyperref}

\usepackage{orcidlink}

\begin{document}
\title{Improving the Transferability of Adversarial Examples by Feature Augmentation}

\titlerunning{Feature Augmentation}

\author{Donghua Wang\inst{1,2} \and
Wen Yao\inst{2,4} \and
Tingsong Jiang \inst{2,4} \and
Xiaohu Zheng \inst{2,4} \and
Junqi Wu \inst{3,4} \and
Xiaoqian Chen \inst{2,4}
}

\authorrunning{Wang et al.}

\institute{Zhejiang University \and
Defense Innovation Institute, Chinese Academy of Military Science \and
Shanghai Jiao Tong Unversity \and
Intelligent Game and Decision Laboratory
}

\maketitle

\begin{abstract}

Despite the success of input transformation-based attacks on boosting adversarial transferability, the performance is unsatisfying due to the ignorance of the discrepancy across models. In this paper, we propose a simple but effective feature augmentation attack (FAUG) method, which improves adversarial transferability without introducing extra computation costs. Specifically, we inject the random noise into the intermediate features of the model to enlarge the diversity of the attack gradient, thereby mitigating the risk of overfitting to the specific model and notably amplifying adversarial transferability. Moreover, our method can be combined with existing gradient attacks to augment their performance further. Extensive experiments conducted on the ImageNet dataset across CNN and transformer models corroborate the efficacy of our method, e.g., we achieve improvement of {\bfseries +26.22\%} and {\bfseries +5.57\%} on input transformation-based attacks and combination methods, respectively.

\end{abstract}

\section{Introduction}
\label{sec:intro}
Deep neural networks (DNNs) have garnered tremendous success in computer vision tasks, such as image classification\cite{he2016deep,chen2021visformer}, object detection \cite{ren2015faster,carion2020end}, and image segmentation\cite{he2017mask,kirillov2023segany}), leading to their widespread deployment in real-world applications. However, DNNs have been proven vulnerable to adversarial examples crafted by imperceptible perturbation but can induce wrong results from DNN models \cite{szegedy2014intriguing,fgsm2015explaining}, which may impose potential risks for DNN-based systems both in the digital \cite{li2023adaptive,wang2024advops} and physical world \cite{athalye2018synthesizing,wang2021dual,wang2022fca,sun2023differential,wang2023rfla}, particularly in security-sensitive domains, such as facial payment, automatic driving, and video surveillance. 


Although existing attacks can effectively incapacitate white-box DNN models, deceiving the black-box model presents a greater challenge due to the practical unavailability of information regarding the black-box model. To address this challenge, a line of work \cite{li2022approximated,tang2023adversarial,sun2023multi,tang2023natural,TANG2024127431,zhao2023minimizing} has emerged, broadly categorized into two-fold: query-based and transfer-based attacks. The former necessitates extensive queries on the back-box model to engender the adversarial example, rendering it time-consuming and costly. Conversely, transfer-based attacks leverage the significant property of adversarial transferability, wherein adversarial examples generated on a substitute model are utilized to target the black-box model, a more practical approach. Accordingly, this study concentrates on transfer-based attacks, with the objective of enhancing adversarial transferability.


Current transfer-based attacks encompass a variety of approaches, including input transformation-based attacks \cite{xie2019improving,dong2019evading,wang2021admix,liang2023styless,wang2023rethinking}, advanced gradient attacks \cite{mim2018BoostingAA,Lin2020Nesterov,wang2021enhancing,li2023adaptive}, ensemble attacks \cite{tramer2018ensemble,xiong2022stochastic,qian2023lea2,chen2024rethinking}, feature-based attacks \cite{ganeshan2019fda,wang2021feature,zhang2022improving,wang2023improving}, etc. Many attacks are resource-intensive and time-consuming; for instance, ensemble attacks necessitate extra memory to load multiple models, while advanced gradient attacks incur additional computation costs to leverage the neighbor gradient to correct the update direction. Among these approaches, the input transformation-based attack is a simple yet efficient method to enhance adversarial transferability. However, they often overlook the model discrepancies, thereby constraining their capability to achieve superior adversarial transferability.

In this work, we propose a simple yet efficient feature augmentation attack aimed at overcoming the overfitting issue by introducing random noise into the intermediate features of the model. Our feature augmentation method enhances adversarial transferability without imposing additional computation costs. Specifically, we first analyze the impact of random noise in features on model performance, revealing the minimal effects on the shallowest and the deepest layers but significant influence on the intermediate layer. Subsequently, we introduce random noise into the shallow layer feature to diversify the attack gradient, effectively circumventing model-specific overfitting and enhancing adversarial transferability. Additionally, we statistically prove that feature augmentation induces a more substantial shift in feature discrepancy than the input transformation-based method. Finally, we integrate our proposed feature augmentation with gradient-based attacks, yielding the feature augmentation (FAUG) attack. Extensive experiments verify the effectiveness of our method. In summary, our main contributions are as follows:


1) Our investigation reveals that the impact of random noise in features varies across different layers of the model, e.g., for CNN models, the shallow and deeper layers exhibit slight effects compared to intermediate layers.


2) We propose a simple yet efficient feature augmentation attack that injects random noise into the model feature to mitigate overfitting and enhance adversarial transferability. Furthermore, our method is compatible with gradient-based attacks, allowing seamless integration.

3) Extensive experiments over CNN and transformer models on the ImageNet dataset validate the effectiveness of the proposed method. Specifically, we gain a notable improvement of {\bfseries +26.22\%} and {\bfseries 5.57\%} in average transferability rate\footnote{We use the average transferability rate to denote the average attack success rate on all black-box models.} compared to the input transformation-based attacks and the advanced gradient attacks, respectively. 

\section{Related work}
\label{sec:related}

\subsection{Adversarial Attacks}
Adversarial attacks can be divided into white-box attacks and black-box attacks. White-box attacks presume that the adversary can access the victim model, thereby allowing them to exploit the model's gradient for devising the attack algorithm. A prominent gradient-based method is the fast gradient-based sign method (FGSM)\cite{fgsm2015explaining}, which generates adversarial examples along the gradient ascent direction of the loss function within a single step. Numerous variants of FGSM have since emerged, including methods such as increasing the iterative step (IFGSM)\cite{ifgsm2018adversarial}, random initialization and fine gradient update (PGD)\cite{pgd2018towards}, applying Nesterov’s accelerated algorithm (NIFSGM)\cite{Lin2020Nesterov}. Other white-box attacks comprise optimized-based attacks (e.g., C\&W attacks\cite{carlini2017towards}) and methods based on generative models (e.g., AdvGAN \cite{xiao2018generating}).

In contrast, the black-box attacks assume that the adversary lacks access to information (e.g., model architecture, training dataset, and parameters) about the victim model. Existing black-box attacks can be divided into query-based and transfer-based methods. Query-based attacks employ evolution algorithms \cite{alzantot2019genattack,andriushchenko2020square,li2022approximated,wang2023rfla} to iteratively search for adversarial examples. However, these attacks typically require enormous queries to the victim model, rendering them impractical due to their computation cost. Transfer-based attacks leverage the adversarial examples generated on the substitute model to attack the victim model, which allows the adversary to perform white-box attacks on the substitute model. Thus, the goal of transfer-based attacks is to enhance attack performance on black-box models. Varying existing approaches include input transformation-based attacks, ensemble attacks, feature-based attacks, and advanced gradient attacks. Input transformation attacks involve techniques such as random resizing and padding of input (DIFGSM\cite{xie2019improving}), convolving the gradient of input images with a Gaussian kernel (TIFGSM\cite{dong2019evading}), multiple image mixture (Admix \cite{wang2021admix}), scaling with multiple factors (SIFGSM\cite{Lin2020Nesterov}) or leveraging frequency domain transformation (SI$^2$FGSM\cite{long2022frequency}). Ensemble attacks combine multiple model outputs  \cite{tramer2018ensemble}, reduce stochastic variance during ensembling of multiple models\cite{xiong2022stochastic}, and exploit the common weakness among different models \cite{chen2024rethinking}. Feature-based attacks devise novel feature-based loss functions \cite {ganeshan2019fda,wang2021feature,zhang2022improving,wang2023improving}. Advanced gradient attacks introduce the momentum to stabilize the gradient updation direction (MIFGSM\cite{mim2018BoostingAA}), variance tuning to correct the gradient(VMIFGSM\cite{wang2021enhancing}), and adaptive momentum variance (AMVIFGSM\cite{li2023adaptive}).

\subsection{Adversarial Defenses}
Concurrently, a range of adversarial defense methods has been proposed, including adversarial training, robust architecture search, preprocessing-based method, denoising-based method, etc. Adversarial training \cite{tramer2018ensemble} is the most efficient defense method, which alternatively generates the adversarial examples and then uses them to train the model. Robust architecture search method \cite{sun2024efficient} aims to find the optimal architecture from the predefined module set through extensive training and evaluation. The preprocessing-based method mitigates potential adversarial perturbation by applying various operations on the input, e.g., JPEG compression \cite{guo2018countering}, random resizing and padding (R\&P)\cite{xie2018mitigating}. Denoising methods purify the input using denoise models to erasure the adversarial perturbation, exemplified by high-level representation guided denoiser (HGD) \cite{liao2018defense}. Other defense methods focus on manipulating model features, such as JPEG-based feature distillation (FD) \cite{liu2019feature} and neural representation purification (NPR) \cite{naseer2020self}. Furthermore, certified defense, such as randomized smoothing(RS) \cite{jia2020certified}, offers reliable defense at specific conditions; combination defense combines different defense methods for stronger defense (NIPS-r3 \footnote{\url{https://github.com/anlthms/nips-2017/tree/master/mmd}}).

\section{Methodology}
\label{sec:method}
In this section, we begin by outlooking the problem statement in Section \ref{subsec:problem}, then devolve into the details of the proposed feature augmentation method in Section \ref{subsec:faug}. Finally, we introduce our attack algorithm incorporating the feature augmentation method in Section \ref{subsec:attack_algorithm}.

\subsection{Problem Statement}
\label{subsec:problem}
Let $(\mathcal{X, Y})$ denote the dataset and the corresponding ground truth label. Given a clean image $x$ sampled from the dataset $\mathcal{X}$, with ground truth label is $y$, a well-trained classification model $f_\theta$ parameterized by $\theta$ performs inference on the image $x$, yielding $f_\theta(x)=\hat{y}$, where $\hat{y}=y$. Our goal is to manufacture an imperceptible adversarial perturbation $\delta$ such that the resulting adversarial examples $x_{adv} = x + \delta$ successfully deceive the model $f_\theta$ i.e., $f_\theta(x_{adv})\neq y$. Therefore, the generation of the adversarial example can be formulated as the following optimization problem:

\begin{equation}
\label{eq:def}
\arg \max_{x_{adv}} \mathcal{L}(f_\theta(x_{adv}), y), ~~~ s.t. ~||\delta||_p \leq \epsilon,
\end{equation}
where $\mathcal{L}(\cdot, \cdot)$ represents the cross-entropy loss in classification task. $||\cdot||_p$ is the $L_p$-norm, and we follow the previous work \cite{dong2019evading,xie2019improving,li2023adaptive} to adopt the $L_\infty$ norm. $\epsilon$ is the maximum allowable modification magnitude of the perturbation $\delta$, constrained within an $L_\infty$ sphere centered at $x$ with a radius of $\epsilon$.

However, the problem defined in Equation \ref{eq:def} can only be solved when the classification model is accessible, as in white-box settings. In scenarios where the model is inaccessible (i.e., black-box settings), solving this problem directly becomes infeasible.  One potential solution to overcome this limitation is to generate the adversarial examples on a substitute (accessible) model $f_\phi$ and then exploit the adversarial transferability to attack the inaccessible model $f_\theta$. Hence, improving the adversarial transferability of adversarial examples generated by the accessible model $f_\phi$ becomes crucial, which is the focus of this paper.

\subsection{Feature Augmentation}
\label{subsec:faug}

\begin{figure}[t]
	\centering
	\begin{minipage}{.45\linewidth}
		\centering
		\includegraphics[width =1.\linewidth]{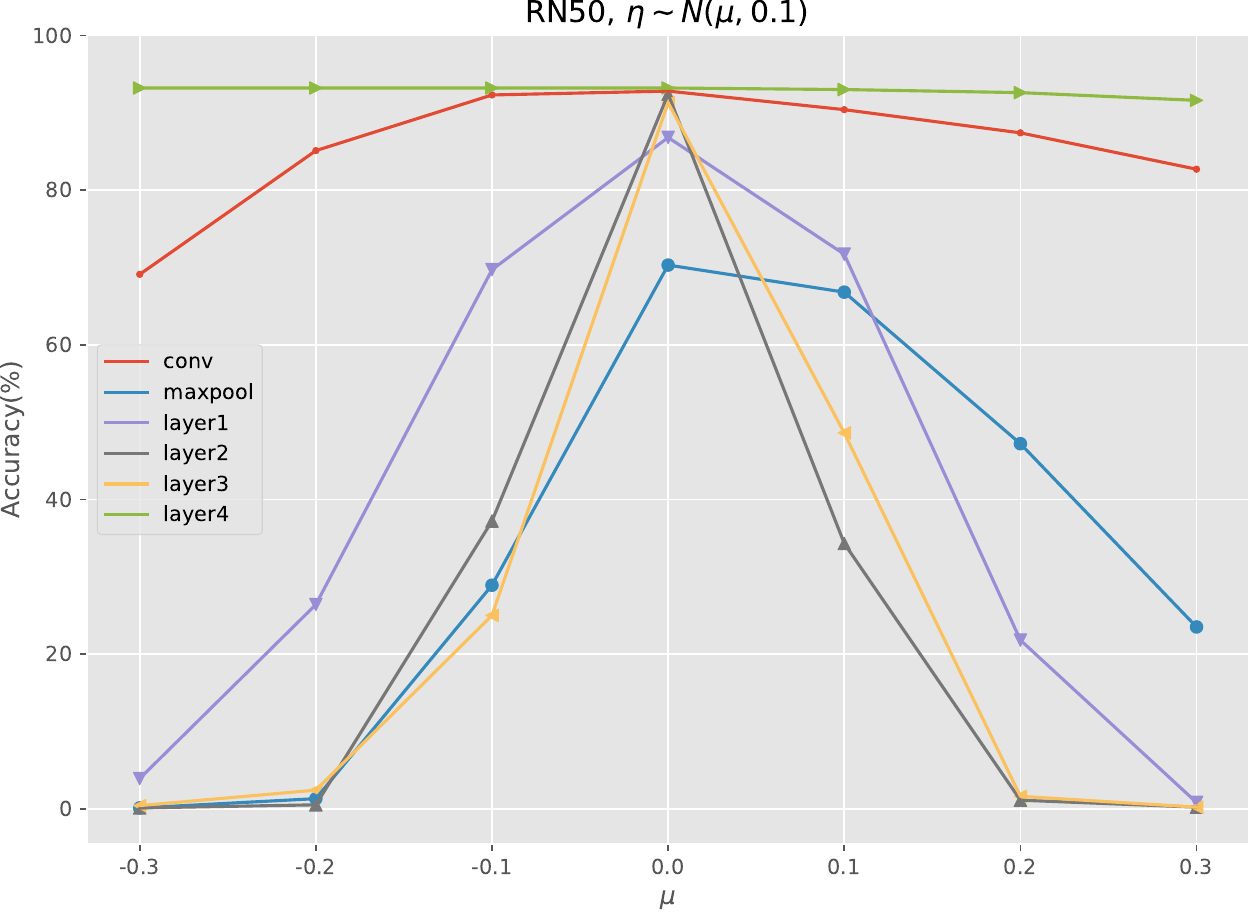}
	\end{minipage}
	\begin{minipage}{.45\linewidth}
		\centering
		\includegraphics[width =1.\linewidth]{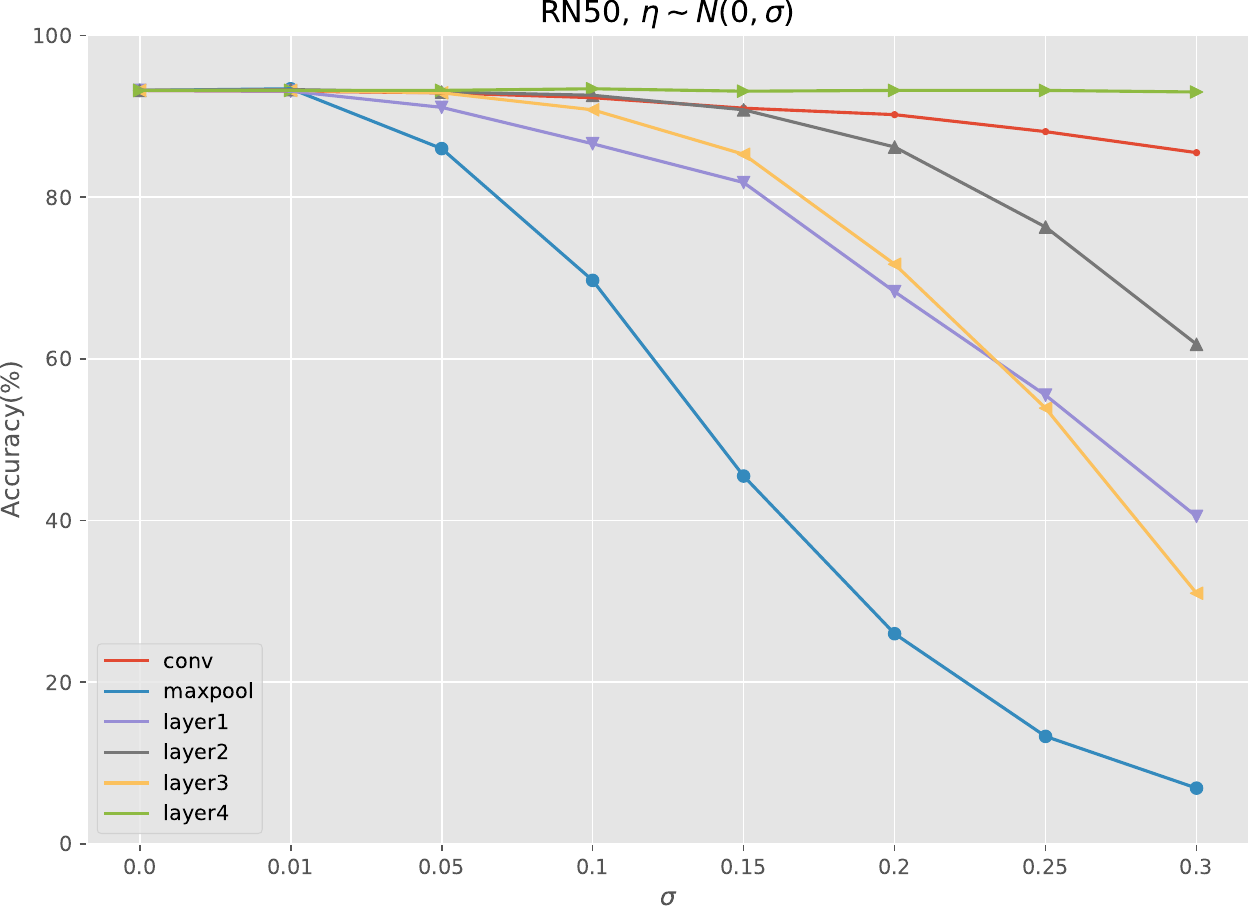}
	\end{minipage}
\caption{Influence of the type of feature augmentation on ResNet50 (abbr.RN50) performance.}
\label{fig:motivation}
\end{figure}

Data augmentation is considered one of the efficient techniques for improving the model's robustness \cite{devries2017cutout,yun2019cutmix,zhang2018mixup,rebuffi2021data} as well as improving the adversarial transferability \cite{xie2019improving,dong2019evading}, owing to its ability to augment the diversity of input data. However, input transformation-based augmentation overlooks the discrepancy between different models, resulting in suboptimal adversarial transferability. In this work, we employ augmentation in the intermediate features to amplify the diversity of the attack gradient, mitigating overfitting and yielding better adversarial transferability. We elucidate the disparity between input transformation and feature augmentation through statistical analysis of the cosine similarity of model logit: (a) between the output logit of the clean image and the corresponding transformation image; and (b) between the output logit of the standard model and the feature-augmented model on clean images. The mean and standard deviation of (a) are 0.9145 and 0.044, while our method (b) yields a lower mean value (0.8971) and higher standard deviation (0.0598), indicating that our method can amplify the diversity of model features and result in superior adversarial transferability.


Formally, given the accessible model $f_\phi$, which is stacked of multiple layers $f_\phi=f^L_\phi(f^{L-1}_\phi(\cdots(f^2_\phi(f^1_\phi(x)))))$, where $f^i_\phi(\cdot)$ denotes the $i$-th layer of the model, $f^1_\phi(\cdot)$ is the input layer, and $f^L_\phi(\cdot)$ is the output layer. We use $f^i_\phi$ to indicate the resulting feature processed by the $f^i_\phi(\cdot)$ layer. To perform the feature augmentation, we inject the random noise $\eta$ into a specific intermediate feature $f^i_\phi$. Mathematically, the augmented feature can be expressed as follows:

\begin{equation}
\hat{f}^i_\phi = f^i_\phi + \eta, ~~~s.t.~~\eta \sim N(\mu, \sigma),
\end{equation}
where $N(\mu, \sigma)$ represents random normal noise with the mean $\mu$ and the standard deviation $\sigma$. Random noise can be seamlessly replaced with other random noise, e.g., uniform noise, which will be elaborated in Section \ref{sec:abla}. Note that the resolution of the injected random noise matches that of the feature $f^i_\phi$. Subsequently, we replace the $f^i_\phi$ with the augmented feature $\hat{f}^i_\phi$ to obtain an augmented model $\hat{f}_\phi$. 

To investigate how the noise strength of feature augmentation affects model performance, we use the augmented model $\hat{f}_\phi$ with different values of $\mu$ and $\sigma$ to classify the dataset, and the evaluation result is illustrated in Figure \ref{fig:motivation}. As we can observe, $\mu$ exerts a greater influence on model accuracy than $\sigma$, likely due to varying $\mu$ significantly impacting the original feature distribution. In contrast, the model feature exhibits a certain robustness to the noise distribution whose $\mu$ is zero and $\sigma$ is small. Therefore, to prevent the significant offset in the feature distribution, we set the $\mu$ to zero and determine $\sigma$  for different models through a grid search, as discussed in Section \ref{sec:abla}.


\subsection{Attack Algorithms}
\label{subsec:attack_algorithm}

Unlike the input transformation-based attacks that necessitate additional processing to augment input image, our feature augmentation attack incurs no extra computation costs apart from adding random noise into the intermediate features. Similar to previous works \cite{xie2019improving,long2022frequency}, our method can be seamlessly integrated with arbitrary gradient-based attacks. In this work, we integrate feature augmentation with MIFGSM \cite{mim2018BoostingAA}, wherein the update of adversarial examples is formulated as follows

\begin{equation}
x^{(0)}_{adv} = x, ~~x^{(t+1)}_{adv}=x^{(t)}_{adv} + \alpha \cdot \text{sign}(g_{t+1}),~ g_{t+1} = \xi \cdot g_{t} + \frac{\nabla_x \mathcal{L}(\hat{f}_\phi(x^{(t)}_{adv}), y)}{||\nabla_x \mathcal{L}(\hat{f}_\phi(x^{(t)}_{adv}), y)||_1}, 
\end{equation}
where $x^{(t)}_{adv}$ and $g_{t}$ represents the generated adversarial examples and the gradient at $t$-th iteration, $\alpha$ denotes the attack step, $\xi$ is the decay factor of the momentum term, $||\cdot||_1$ denotes the $L_1$-norm. The detailed algorithm is described in Algorithm \ref{alg:algorithm}.

\begin{algorithm}[t]
	\caption{Feature Augmentation Attacks (FAUG)}
	\label{alg:algorithm}
	\textbf{Input}: A feature augmented model $\hat{f}_\phi$, a clean image $x$, the perturbation bound $\epsilon$, the attack step $\alpha$, the decay factor $\xi$, the number of iterative step T \\
	\textbf{Output}: The adversarial example $x_{adv}$
	\begin{algorithmic}[1] 
		\STATE $x^{(0)}_{adv} = x$, $g_0=0$
		\FOR {$t=0,...,T-1$}
		\STATE  Input $x^{(t)}_{adv}$ to $\hat{f}_\phi$ and obtain the gradient $\nabla_x \mathcal{L}(\hat{f}_\phi(x^{(t)}_{adv}), y)$
		\STATE  Update the accumulated gradient $g_{t+1} = \xi \cdot g_{t} + \frac{\nabla_x \mathcal{L}(\hat{f}_\phi(x^{(t)}_{adv}), y)}{||\nabla_x \mathcal{L}(\hat{f}_\phi(x^{(t)}_{adv}), y)||_1}$
		\STATE Update adversarial examples $x^{(t+1)}_{adv}=x^{(t)}_{adv} + \alpha \cdot \text{sign}(g_{t+1})$
		\STATE Obtain the adversarial perturbation $\delta = \min( \max(x^{(t+1)}_{adv}-x, \epsilon), -\epsilon)$
		\STATE Update the adversarial examples $x^{(t+1)}_{adv} =  \min( \max(x + \delta, 1), 0)$
		\ENDFOR
	\RETURN $x_{adv} = x^{(T)}_{adv}$
	\end{algorithmic}
\end{algorithm}

\section{Experiment}
\label{sec:exp}
\subsection{Experiment Settings}

\indent{\bfseries Dataset.} To evaluate the effectiveness of the proposed method, we follow previous works \cite{dong2019evading,li2023adaptive} to adopt the commonly used ImageNet-compatible dataset\footnote{\url{https://github.com/cleverhans-lab/cleverhans/tree/master/cleverhans_v3.1.0/examples/nips17_adversarial_competition/dataset}}, which contains 1000 images and the resolution is adjusted to 3$\times$224$\times$224.

{\bfseries Evaluation Models.} We choose eight standardly trained models provided by \textsf{timm} package \cite{rw2019timm} to evaluate our method, consisting of four CNN-based models: ResNet50 (RN50)\cite{he2016deep}, DenseNet121 (DN121)\cite{huang2017densely}, VGG19BN \cite{simonyan2014very}, and ResNeXt50\_32x4d (RNX50)\cite{xie2017aggregated} as well as four transformer-based models, including Visformer-S\cite{chen2021visformer}, ViT-B/16\cite{dosovitskiy2021an}, PiT-B\cite{heo2021rethinking}, and Swin-B/S3\cite{liu2021swin}. To evaluate our method under defense method, we choose seven defense methods, including two adversarially trained models comprised of AdvIncV3, AdvIncResV2$_{ens}$ \cite{tramer2018ensemble}, R\&P\cite{xie2018mitigating}, NIPS-r3, JPEG\cite{guo2018countering}, RS\cite{jia2020certified}, and NRP\cite{naseer2020self}. 

{\bfseries Comparison Methods.} We compare our method with five widely used attacks, including MIFGSM\cite{mim2018BoostingAA}, DIFGSM\cite{xie2019improving}, TIFGSM\cite{dong2019evading}, NIFSGM\cite{Lin2020Nesterov}. SI$^2$FGSM\cite{long2022frequency}. Moreover, we also integrate our method into the advanced gradient attacks to investigate whether our method can enhance their performance, including SINIFGSM \cite{Lin2020Nesterov}, VNIFGSM\cite{wang2021enhancing}, VMIFGSM\cite{wang2021enhancing}, AMVIFSGM \cite{li2023adaptive}.  

{\bfseries Implementation Details.}  We follow previous works \cite{xie2019improving,dong2019evading,wang2021enhancing} to set the maximum allowable modification $\epsilon=16$, the number of iteration T = 10, $\alpha$ to 2/255, $\xi$ to 1.0. The detailed selection of layer and $\sigma$ values for different models are provided in Supplementary Sec. \ref{supp:details}. To implement the comparison attacks, we utilize the attack toolbox \textsf{torchattacks} \cite{kim2020torchattacks} while remaining default parameters except for the epsilon $\epsilon$ and iteration step T. Moreover, we utilize the interface provided by \textsf{torchattacks} to reproduce SI$^2$FGSM and AMVIFGSM, with parameter settings consistent with other available attacks. All codes are implemented in PyTorch\footnote{Code will be released after being accepted.}  and executed on a single NVIDIA Tesla V100 GPU.


\begin{table}[t]
\centering
\tiny
\setlength\tabcolsep{2pt}
\caption{Comparison results of various attack methods in terms of attack success rate (\%). The bold item indicates the best one. Item with $\star$ superscript is white-box attacks, and the others is black-box attacks. AVG column indicates the average attack success rate on black-box models.}
\label{tab:compare_result}
\begin{tabular}{ccccccccccc}
\hline
                 & Method     & RN50          & DN121         & VGG19BN     & RNX50            & Visformer-S   & PiT-B         & ViT-B/16      & Swin-B/S3     & AVG            \\ \hline
RN50             & MIFGSM     & 100$^\star$   & 88.5          & 79.0          & 90.0             & 41.5          & 31.8          & 31.7          & 26.5          & 55.57          \\
                 & DIFGSM     & 100$^\star$   & 90.0          & 85.0          & 92.5             & 42.0          & 27.9          & 21.8          & 23.7          & 54.70          \\
                 & TIFGSM     & 100$^\star$   & 78.1          & 66.7          & 78.8             & 23.3          & 14.2          & 30.8          & 11.5          & 43.34          \\
                 & NIFGSM     & 100$^\star$   & \textbf{90.5} & 85.5          & 92.6             & 42.6          & 31.5          & 33.6          & 25.2          & 57.36          \\
                 & SI$^2$FGSM & 100$^\star$   & 81.5          & 81.2          & 85.1             & 31.0          & 20.7          & 22.9          & 20.8          & 49.03          \\
                 & FAUG       & 100$^\star$   & 90.4          & \textbf{86.8} & \textbf{92.8}    & \textbf{50.4} & \textbf{38.5} & \textbf{38.3} & \textbf{34.3} & \textbf{61.64} \\ \hline
DN121            & MIFGSM     & 86.8          & 100$^\star$   & 81.7          & 84.6             & 49.3          & 34.7          & 37.2          & 29.7          & 57.71          \\
                 & DIFGSM     & 84.6          & 100$^\star$   & 81.2          & 84.3             & 42.8          & 24.6          & 23.9          & 21.1          & 51.79          \\
                 & TIFGSM     & 68.6          & 100$^\star$   & 64.4          & 70.4             & 26.6          & 13.9          & 33.4          & 11.7          & 41.29          \\
                 & NIFGSM     & 90.5          & 100$^\star$   & 87.7          & 89.9             & 53.3          & 34.3          & 38.2          & 29.9          & 60.54          \\
                 & SI$^2$FGSM & 89.9          & 100$^\star$   & 87.1          & 87.9             & 49.2          & 29.3          & 26.6          & 26.6          & 56.66          \\
                 & FAUG    & \textbf{91.9}    & 100$^\star$   & \textbf{89.6} & \textbf{91.0}    & \textbf{64.7} & \textbf{46.8} & \textbf{48}   & \textbf{40.6} & \textbf{67.51} \\ \hline
VGG19BN         & MIFGSM      & 67.7          & 71.0          & 100$^\star$   & 61.2              & 37.1          & 24.1          & 26.2          & 21.1          & 44.06          \\
                 & DIFGSM     & 64.3          & 67.8          & 100$^\star$   & 60.8             & 30.5          & 17.5          & 15.3          & 15.2          & 38.77          \\
                 & TIFGSM     & 51.6          & 60.5          & 100$^\star$   & 55.5             & 20.9          & 11.8          & 26.4          & 9.9           & 33.80          \\
                 & NIFGSM     & 71.8          & 73.8          & 100$^\star$   & 64.0             & 37.7          & 24.2          & 28.3          & 22.3          & 46.01          \\
                 & SI$^2$FGSM  & 61.3          & 59.4          & 99.9$^\star$ & 49.0             & 22.4          & 15.0          & 18.8          & 14.2          & 34.30          \\
                 & FAUG    & \textbf{82.8} & \textbf{83.9}    & 100$^\star$   & \textbf{75.8}    & \textbf{47.0}   & \textbf{32.1} & \textbf{36.2} & \textbf{28.2} & \textbf{55.14} \\ \hline
RNX50            & MIFGSM     & 84.0          & 80.1          & 70.7          & 100$^\star$      & 41.1          & 30.4          & 30.5          & 26.1          & 51.84          \\
                 & DIFGSM     & 88.3          & 84.5          & 76.0          & 99.9$^\star$     & 41.0          & 25.7          & 20.7          & 20.5          & 50.96          \\
                 & TIFGSM     & 69.8          & 69.1          & 57.4          & 99.8$^\star$     & 24.2          & 14.6          & 29.4          & 11.1          & 39.37          \\
                 & NIFGSM     & 89.0          & 83.3          & 78.9          & 100$^\star$      & 43.3          & 30.4          & 32.1          & 26.1          & 54.73          \\
                 & SI$^2$FGSM & 87.9          & 81.1          & 77.0          & 99.9$^\star$     & 35.3          & 24.0          & 23.2          & 22.1          & 50.09          \\
                 & FAUG    & \textbf{90.2} & \textbf{84.9} & \textbf{78.9} 	  & 100$^\star$      & \textbf{51.9} & \textbf{38.9} & \textbf{36.6} & \textbf{32.9} & \textbf{59.19} \\  \hline
Visformer-S      & MIFGSM     & 67.0          & 69.4          & 72.0          & 66.7             & 100$^\star$   & 66.3          & 44.4          & 60.7          & 63.79          \\
                 & DIFGSM     & 65.4          & 68.7          & 72.4          & 68.9             & 99.6$^\star$  & \textbf{71.5} & 31.7          & 64.8          & 63.34          \\
                 & TIFGSM     & 50.7          & 57.3          & 55.0          & 57.6             & 98.6$^\star$  & 49.1          & 44.1          & 38.1          & 50.27          \\
                 & NIFGSM     & 70.0          & 74.8          & 76.8          & 70.1             & 100$^\star$   & 68.0          & 45.2          & 63.0          & 66.84          \\
                 & SI$^2$FGSM & 71.5          & 76.8          & 76.5          & 69.8             & 98.8$^\star$  & 71.1          & 45.3          & \textbf{69.0} & 68.57          \\
                 & FAUG    & \textbf{75.8} & \textbf{80.4} & \textbf{83.3} & \textbf{75.0}       & 100$^\star$   & 70.4          & \textbf{46.6} & 65.9          & \textbf{71.06} \\ \hline
PiT-B            & MIFGSM     & 61.9          & 59.9          & 69.0          & 60.1             & 60.6          & 100$^\star$   & 40.7          & 51.7          & 57.70          \\
                 & DIFGSM     & 55.8          & 55.9          & 60.7          & 58.2             & \textbf{72.9} & 99.8$^\star$  & 34.4          & \textbf{65.0} & 57.56          \\
                 & TIFGSM     & 39.0          & 44.9          & 43.7          & 44.8             & 54.7          & 97.3$^\star$  & 40.4          & 40.1          & 43.94          \\
                 & NIFGSM     & 63.4          & 62.7          & 71.1          & 62.3             & 62.7          & 100$^\star$   & 41.4          & 54.9          & 59.79          \\
                 & SI$^2$FGSM & 57.3          & 58.2          & 62.1          & 56.8             & 68.9          & 97.9$^\star$  & 40.9          & 61.5          & 57.96          \\
                 & FAUG       & \textbf{65.8} & \textbf{64.9} & \textbf{72.2} & \textbf{63.7}    & 69.7          & 100$^\star$   & \textbf{44.9} & 59.8          & \textbf{63.00} \\ \hline
ViT-B/16         & MIFGSM     & 51.4          & 52.7          & 56.7          & 46.7             & 38.7          & 33.8          & 100$^\star$   & 36.3          & 45.19          \\
                 & DIFGSM     & 33.7          & 37.4          & 36.9          & 32.4             & 27.9          & 27.2          & 100$^\star$   & 26.2          & 31.67          \\
                 & TIFGSM     & 28.1          & 32.4          & 29.5          & 29.9             & 18.3          & 16.6          & 99.6$^\star$  & 14.1          & 24.13          \\
                 & NIFGSM     & 52.6          & \textbf{55.8} & 59.7		  & \textbf{50.3}    & 39.5          & 35.3          & 100$^\star$   & 35.8          & 47.00 		 \\
                 & SI$^2$FGSM & 25.7          & 29.2          & 32.8          & 23.5             & 15.4          & 13.5          & 100$^\star$   & 14.5          & 22.09          \\
                 & FAUG       & 52.6          & 55.1          & 59.7		  & 48.0             & \textbf{41.2} & \textbf{38.3} & 100$^\star$   & \textbf{40.7} & \textbf{47.94}  \\ \hline
Swin-B/S3        & MIFGSM     & 44.9          & 45.0          & 52.2          & 40.3             & 48.0          & 40.5          & 32.0          & 99.8$^\star$  & 43.27          \\
                 & DIFGSM     & 47.9 		  & 50.3 		  & 57.0 		  & 48.5    		 & \textbf{64.3} & \textbf{63.6} & 29.6          & 96.8$^\star$  & 51.60         \\
                 & TIFGSM     & 37.5          & 44.7          & 41.6          & 43.2             & 50.6          & 47.7          & 41.9          & 85.5$^\star$   & 43.89          \\
                 & NIFGSM     & 44.5          & 44.8          & 52.0          & 38.2             & 45.8          & 42.0          & 32.0          & \textbf{99.9}$^\star$   & 42.76          \\
                 & SI$^2$FGSM & 45.6          & 46.0          & 48.1          & 43.2             & 55.4          & 49.5          & 38.2          & 92.1$^\star$  & 46.57          \\
                 & FAUG       & \textbf{54.2} & \textbf{53.2} & \textbf{57.3} & \textbf{50.0}    & 60.1          & 56.1          & \textbf{48.2} & 88.4$^\star$  & \textbf{54.16} \\ \hline
\end{tabular}
\end{table}

\subsection{Main results}
In this section, we compare the proposed FAUG and existing attack methods to verify its effectiveness. Comparison method includes classical attacks (i.e., MIFGSM \cite{mim2018BoostingAA} and NIFGSM \cite{Lin2020Nesterov}) and input transformation-based attacks (i.e., DIFGSM \cite{xie2019improving}, TIFGSM \cite{dong2019evading} and SI$^2$FGSM \cite{long2022frequency}). Note that our FAUG is the attack that applies MIFGSM on the feature-augmented model, while the comparison methods target the standard trained model. Table \ref{tab:compare_result} lists the comparison results across eight models.

\begin{table}[htbp]
\centering
\tiny
\setlength\tabcolsep{2pt}
\caption{Evaluation results of the advanced attacks combined with feature augmentation model in terms of attack success rate (\%). The bold item indicates the best one. Item with $\star$ superscript is white-box attacks, and the others is black-box attacks. AVG column indicates the average attack success rate on black-box models.}
\label{tab:incremental_attack}
\begin{tabular}{c|c|ccccccccc}
\hline
            & Model         & RN50          & DN121         & VGG19BN     & RNX50         & Visformer-S   & PiT-B         & ViT-B/16      & Swin-B/S3     & AVG            \\ \hline
\multirow{7}{*}{RN50}        
			& VNIFGSM       & 100$^\star$    & \textbf{94.9} & \textbf{94.1} & 96.3          & \textbf{69.5} & 53.0          & 46.5          & 44.2          & 71.21          \\ 
            & VNIFGSM-FAUG  & 100$^\star$   & 94.2          & 93.3          & 96.3          & 69.4          & \textbf{54.2} & \textbf{48.8} & \textbf{47.9} & \textbf{72.01} \\  \cline{2-11}
            & VMIFGSM       & \textbf{100}$^\star$  & 93.0          & 91.5          & \textbf{95.3} & 64.8          & 48.7          & 43.6          & 45.1          & 68.86          \\
            & VMIFGSM-FAUG  & 99.9$^\star$  & \textbf{93.2} & \textbf{91.9} & 95.1          & \textbf{67.7} & \textbf{52.2} & \textbf{48.6} & \textbf{46.3} & \textbf{70.71} \\  \cline{2-11}
            & SINIFGSM      & 100$^\star$   & 95.7          & 92.5          & \textbf{96.6} & 61.2          & 43.5          & 42.6          & 36.4          & 66.93          \\ 
            & SINIFGSM-FAUG & 100$^\star$   & \textbf{96.0} & \textbf{93.8} & 96.3          & \textbf{63.1} & \textbf{44.5} & \textbf{45.1} & \textbf{39.2} & \textbf{68.29} \\  \cline{2-11}
            & AMVIFGSM      & 100$^\star$   & 95.7          & 93.5          & 97.0          & 66.5          & 51.8          & 54.1 		    & 47.2 			& 72.26   \\
            & AMVIFGSM-FAUG & 100$^\star$   & \textbf{95.8} & \textbf{94.5} & \textbf{97.5} & \textbf{70.6} & \textbf{54.8} & \textbf{56.0} & \textbf{48.3} & \textbf{73.93}   \\ \hline
\multirow{7}{*}{DN121}       
			& VNIFGSM       & \textbf{94.9} & 100$^\star$     & 92.5          & \textbf{94.2} & 76.4          & 54.2          & 54.0          & 49.4          & 73.66          \\
            & VNIFGSM-FAUG  & 94.5          & 100$^\star$     & \textbf{92.9} & 93.5          & \textbf{77.7} & \textbf{58.9} & \textbf{57.6} & \textbf{54.0}  & \textbf{75.59} \\  \cline{2-11}
            & VMIFGSM       & 93.0          & 100$^\star$     & 90.4          & 92.0          & 72.3          & 51.7          & 51.1          & 47.7          & 71.17          \\
            & VMIFGSM-FAUG  & \textbf{94.2} & 100$^\star$     & \textbf{91.9} & \textbf{93.2} & \textbf{75.4} & \textbf{57.4} & \textbf{56.4} & \textbf{51.9} & \textbf{74.34} \\  \cline{2-11}
            & SINIFGSM      & 94.4          & 100$^\star$     & \textbf{93.2} & \textbf{95.0}  & 68.8          & \textbf{47.5} & 50.7          & 39.9          & 69.93          \\
            & SINIFGSM-FAUG & \textbf{94.7} & 100$^\star$     & 92.6          & 94.4          & \textbf{70.3} & 46.6          & \textbf{52.7} & \textbf{42.6} & \textbf{70.56} \\   \cline{2-11}
            & AMVIFGSM      & \textbf{95.5} & 100$^\star$     & 93.4          & 94.5          & 77.0          & \textbf{59.8} & 60.4          & 53.5          & 76.30          \\
            & AMVIFGSM-FAUG & 95.1          & 100$^\star$     & \textbf{94.1} & \textbf{95.1} & \textbf{77.5} & 59.2          & \textbf{60.9} & \textbf{53.9} & \textbf{76.54} \\  \hline
\multirow{7}{*}{VGG19BN}   & VNIFGSM       & 87.5             & 88.3          & 100$^\star$   & 84.3          & 61.3          & 43.8          & \textbf{41.2} & \textbf{39.2} & 63.66          \\
            & VNIFGSM-FAUG  & \textbf{87.8} & \textbf{88.9} & 100$^\star$    & \textbf{85.0} & \textbf{62.6} & \textbf{44.5} & 40.8          & 38.6          & \textbf{64.03} \\  \cline{2-11}
            & VMIFGSM       & 84.5          & 86.4          & 100$^\star$    & 81.0          & 58.4          & 42.1          & 41.3          & 35.7          & 61.34          \\
            & VMIFGSM-FAUG  & \textbf{85.9} & \textbf{87.0} & 100$^\star$    & \textbf{83.1} & \textbf{60.2} & \textbf{43.7} & \textbf{42.1} & \textbf{39.2} & \textbf{63.03} \\  \cline{2-11}
            & SINIFGSM      & 86.4          & \textbf{91.2} & 100$^\star$    & 82.7          & 56.5          & 35.9          & 37.0          & 30.5          & 60.03          \\
            & SINIFGSM-FAUG & \textbf{89.3} & 90.9          & 100$^\star$    & \textbf{85.6} & \textbf{62.5} & \textbf{39.6} & \textbf{41.4} & \textbf{32.1} & \textbf{63.06} \\   \cline{2-11}
            & AMVIFGSM      & 89.0          & 89.1          & 100$^\star$    & 85.9          & \textbf{58.1} & 39.2          & 46.5          & 35.0          & 63.26          \\
            & AMVIFGSM-FAUG & \textbf{89.8} & \textbf{89.7} & 100$^\star$    & \textbf{87.4} & 57.8          & \textbf{39.5} & \textbf{46.8} & \textbf{36.2} & \textbf{63.89} \\ \hline
\multirow{7}{*}{RNX50}       
			& VNIFGSM       & \textbf{94.6} & \textbf{91.9} & 88.6          & 100$^\star$     & 68.8          & 52.9          & 46.7          & 45.8          & 69.90          \\
            & VNIFGSM-FAUG  & 94.1          & 91.8          & \textbf{88.8} & 100$^\star$    & \textbf{69.9} & \textbf{55.0} & \textbf{49.7} & \textbf{46.9} & \textbf{70.89} \\  \cline{2-11}
            & VMIFGSM       & 92.2          & 89.2          & 84.9          & 100$^\star$    & 65.2          & 49.5          & 44.0          & 42.5          & 66.79          \\
            & VMIFGSM-FAUG  & \textbf{93.1} & \textbf{91.5} & \textbf{86.4} & 100$^\star$    & \textbf{67.2} & \textbf{52.7} & \textbf{47.8} & \textbf{46.6} & \textbf{69.33} \\  \cline{2-11}
            & SINIFGSM      & 94.6          & 93.6          & 89.0          & 100$^\star$    & 60.2          & 42.3          & 41.8          & 34.7          & 65.17          \\
            & SINIFGSM-FAUG & \textbf{95.5} & \textbf{94.9} & \textbf{91.5} & 100$^\star$    & \textbf{64.6} & \textbf{47.0} & \textbf{48.1} & \textbf{38.7} & \textbf{68.61} \\  \cline{2-11}
            & AMVIFGSM      & 94.2          & 92.1          & 89.0          & \textbf{100}$^\star$    & \textbf{68.9} & \textbf{55.5} & 54.0   & 48.0         & 71.67          \\
            & AMVIFGSM-FAUG & \textbf{96.1} & \textbf{92.9} & \textbf{90.6} & 99.9$^\star$   & 67.9          & 53.8          & \textbf{54.9} & 48.0          & 72.03          \\  \hline
\multirow{7}{*}{Visformer-S} 
			& VNIFGSM       & 85.1          & 88.0          & 88.4          & 87.4          & \textbf{100}$^\star$    & 87.0 & 65.9          & \textbf{86.0} & 83.97          \\
            & VNIFGSM-FAUG  & \textbf{89.9} & \textbf{91.5} & \textbf{91.3} & \textbf{89.3} & 99.9$^\star$   & \textbf{87.6} & \textbf{69.1} & 85.8          & \textbf{86.36} \\  \cline{2-11}
            & VMIFGSM       & 81.1          & 85.1          & 84.9          & 82.5          & 100$^\star$    & 84.3          & 61.1          & 81.9          & 80.13          \\
            & VMIFGSM-FAUG  & \textbf{87.1} & \textbf{90.8} & \textbf{90.1} & \textbf{88.1} & 100$^\star$    & \textbf{86.7} & \textbf{67.5} & \textbf{82.9} & \textbf{84.74} \\  \cline{2-11}
            & SINIFGSM      & 79.1          & 83.9          & 84.2          & 81.2          & 100$^\star$    & 80.0          & 56.0          & 75.9          & 77.19          \\
            & SINIFGSM-FAUG & \textbf{85.9} & \textbf{89.4} & \textbf{90.1} & \textbf{86.4} & 100$^\star$    & \textbf{82.3} & \textbf{60.1} & \textbf{77.9} & \textbf{81.73} \\  \cline{2-11}
            & AMVIFGSM      & 84            & 85.6          & 87.2          & 85.2          & \textbf{99.9}$^\star$   & \textbf{87.5} & 71.8          & \textbf{84.9} & 83.74          \\
            & AMVIFGSM-FAUG & \textbf{88.4} & \textbf{89.1} & \textbf{89.4} & \textbf{86.2} & 99.2$^\star$   & 87.1          & \textbf{72.9} & 84.4          & \textbf{85.36} \\  \hline
\multirow{7}{*}{PiT-B}
	        & VNIFGSM       & 75.2          & 75.5          & 78.0          & 75.6          & 80.9          & 100$^\star$    & 56.7          & 75.8          & 73.96          \\
            & VNIFGSM-FAUG  & \textbf{78.1} & \textbf{78.5} & \textbf{80.3} & \textbf{77.5} & \textbf{85.6} & 100$^\star$    & \textbf{61.7} & \textbf{81.5} & \textbf{77.60} \\ \cline{2-11}
            & VMIFGSM       & 71.3          & 70.3          & 75.4          & 71.7          & 78.4          & 100$^\star$    & 54.3          & 72.4          & 70.54          \\
            & VMIFGSM-FAUG  & \textbf{75.9} & \textbf{76.4} & \textbf{79.0}   & \textbf{76.7} & \textbf{83.3} & 100$^\star$    & \textbf{60.4} & \textbf{81.1} & \textbf{76.11} \\ \cline{2-11}
            & SINIFGSM      & 72.5          & 72.3          & 77.7          & 71.6          & 75.9          & 100$^\star$    & 50.6          & 65.9          & 69.50          \\
            & SINIFGSM-FAUG & \textbf{76.4} & \textbf{76.9} & \textbf{80.2} & \textbf{76.2} & \textbf{83.0} & 100$^\star$    & \textbf{56.2} & \textbf{75.8} & \textbf{74.96} \\  \cline{2-11}
            & AMVIFGSM      & 76.9          & 77.1          & 77.2          & 75.9          & 82.9          & \textbf{100}$^\star$    & 63.8          & 79.7          & 76.21          \\
            & AMVIFGSM-FAUG & \textbf{79.3} & \textbf{79.9} & \textbf{80.1} & \textbf{79.9} & \textbf{84.3} & 99.7$^\star$   & \textbf{68.4} & \textbf{83.2} & \textbf{79.30} \\  \hline
\multirow{7}{*}{ViT-B/16}   
			& VNIFGSM       & 58.7          & 59.7			& \textbf{65.0} & 55.4			 & \textbf{48.3} & \textbf{43.0} & 100$^\star$      & 46.5          & 53.80			 \\
            & VNIFGSM-FAUG  & \textbf{59.8} & 59.7          & 64.5          & \textbf{55.8}  & 46.9    		 & 42.5          & 100$^\star$      & \textbf{48.5}& \textbf{53.96}          \\  \cline{2-11}
            & VMIFGSM       & 53.4          & 54.0          & \textbf{59.7} & \textbf{49.4} & \textbf{43.8} & 39.8           & 100$^\star$   & 44.3          & 49.20          \\
            & VMIFGSM-FAUG  & \textbf{54.7} & \textbf{55.0} & 59.4          & 49.3          & 43.3          & \textbf{40.2}  & 100$^\star$   & \textbf{44.9} & \textbf{49.54} \\ \cline{2-11}
            & SINIFGSM      & 56.4			& 61.5          & 63.5			& \textbf{55.1} & 43.8			 & \textbf{36.1} & 100$^\star$    & 40.0       & 50.91 \\
            & SINIFGSM-FAUG & \textbf{57.0} & \textbf{62.3} & \textbf{65.4} & 53.3          & \textbf{44.3}  & 36.5			 & 100$^\star$    & \textbf{40.3} & \textbf{51.14}          \\ \cline{2-11}
            & AMVIFGSM      & 54.8 			& 55.3			 & 61.0  	    & 50.7          & 43.4           & 38.1			 & 100$^\star$    & 43.8 & 49.59 \\
            & AMVIFGSM-FAUG & 54.8          & \textbf{58.4}  & \textbf{61.9}  & \textbf{52.2} & \textbf{43.9} & \textbf{38.9}& 100$^\star$  & \textbf{44.3}   & \textbf{50.63}          \\  \hline
\multirow{7}{*}{Swin-B/S3}   
			& VNIFGSM       & 57.8          & 58.0          & 60.4          & 55.8          & 67.8          & 66.5          & 50.5          & \textbf{99.6}$^\star$ & 59.54          \\
            & VNIFGSM-FAUG  & \textbf{58.9} & \textbf{59.8} & \textbf{61.7} & \textbf{57.7} & \textbf{70.2} & \textbf{67.6} & \textbf{52.2} & 96.2$^\star$          & \textbf{61.16} \\ \cline{2-11}
            & VMIFGSM       & 56.7          & \textbf{58.9} & 59.9          & 55.7          & 67.7          & 66.1          & 50.0          & \textbf{98.9}$^\star$ & 59.29          \\
            & VMIFGSM-FAUG  & \textbf{58.6} & 58.0          & \textbf{62.2} & \textbf{56.2} & \textbf{69.2} & \textbf{67.2} & \textbf{51.9} & 94.7$^\star$          & \textbf{60.47} \\   \cline{2-11}
            & SINIFGSM      & 52.8          & 53.5          & 59.6          & 50.3          & \textbf{60.9} & 52.8          & 39.5          & \textbf{99.9}$^\star$ & 52.77          \\
            & SINIFGSM-FAUG & \textbf{55.1} & \textbf{55.5} & \textbf{62.0} & \textbf{52.6} & 59.9          & \textbf{56.1} & \textbf{43.9} & 99.5$^\star$          & \textbf{55.01} \\   \cline{2-11}
            & AMVIFGSM      & 63.6          & 62.8          & 65.5          & 64.0          & \textbf{73.9} & 72.1          & 60.0          & \textbf{99.1}$^\star$ & 65.99          \\
            & AMVIFGSM-FAUG & \textbf{65.0} & \textbf{65.0} & \textbf{66.0} & \textbf{65.3} & 72.4          & \textbf{73.9} & \textbf{60.9} & 94.3$^\star$          & \textbf{66.93} \\ \hline
\end{tabular}
\end{table}

From Table \ref{tab:compare_result}, it is evident that our FAUG consistently outperforms comparison methods on all black-box models. More specifically, we achieve an average transferability rate of 59.96\% overall on eight different models, significantly surpassing MI-, DI-, TI-, NI- and SI$^2$FGSM, which stands at 52.39\%, 50.05\%, 40\%, 54.38\%, 48.16\%. Notably, our method exhibits the largest improvement of {\bfseries 26.22\%} in the case of TIFGSM against DN121, in which we obtained an average transferability rate of 67.51\% compared to TIFGSM's rate of 41.29\%. Moreover, we observe that Visformer-S is the most vulnerable model, with an average transferability rate of 63.98\% under six attacks, which breaks the impression that the transformer model is more robust than CNN models \cite{bhojanapalli2021understanding}. However, it is also noteworthy that the most robust model against adversarial attacks is a transformer model (i.e., ViT-B/16) as well, with an average transferability rate of 36.34\%. These analyses indicate that model robustness is more closely associated with architecture design rather than specific modules (e.g., convolutional or attention layer). Finally, we acknowledge that our method falls behind the comparison method in some cases, both in white-box and black-box settings, possibly due to randomness introduced by the random noise in feature augmentation.

\subsection{Combined with Advanced Gradient Attacks}
 
In this section, we explore the potential enhancement of attack performance by integrating our method with advanced gradient attacks. Specifically, we integrate our feature augmentation method by injecting the random noise into the specific intermediate feature of the standard model,  followed by the execution of advanced gradient attacks on such a modified model. The combined methods are denoted with postfix "-FAUG", e.g., VNIFGSM-FAUG. Table \ref{tab:incremental_attack} reports the evaluation results on eight different models. 

From Table \ref{tab:incremental_attack}, we can draw the following conclusions: on the one hand, our method consistently enhances the performance of original attacks in terms of average transferability rate. Specifically, we achieve an average improvement in transferability rate of 1.98\%, with the maximum improvement reaching {\bfseries 5.57\%} when using VMIFGSM-FAUG on PiT-B. On the other hand, different attacks exhibit varying results across different models. For example, among eight models, the largest disparity in attack performance improvement is 3.08\% observed on RNX50, in which SINIFGSM-FAUG results in 3.44\% while AMVIFGSM-FAUG results in 0.36\%. Conversely, the smallest disparity in performance improvement of 0.89\% is observed in  ViT-B/16. Furthermore, among the four attacks, the largest improvement in attack performance of 5.23\% is observed on VMIFGSM-FAUG and SINIFGSM-FAUG when targeting PiT-B (best) and ViT-B/16 (worst). In contrast, AMVIFGSM-FAUG exhibits the smallest disparity in performance improvement, which is 2.84\% between PiT-B (3.09\%) and DN121 (0.24\%). The discrepancy may be attributed to the smaller improvement space of AMVIFGS compared to other attacks.

\subsection{Evaluation on Adversarial Defense Method}

To further explore the effectiveness of the proposed method under adversarial defense, we conducted experiments by applying seven common defense methods to adversarial examples generated by RN50. The evaluation results are reported in Table \ref{tab:defense}. As we can see, on the one side, our method outperforms the comparison method under adversarial defense in terms of average attack success rate. Specifically, we achieve an average success rate of 58.99\% across seven defenses, obtaining an improvement of 2.08\%$\sim$5.45\% compared to the comparison method (i.e., MI-, DI-, TI-, NI-, SI$^2$-FGSM). On the other side, our combination version with advanced gradient attacks also exhibits superior performance to the vanilla approach under defense, and we observe improvements ranging from 0.39\% to 2.7\%. However, we observe that our method does not consistently achieve the best result in all cases. We speculate that the possible reason is the discrepancy in the underlying defense mechanism among different defense methods. Nevertheless, we consistently achieve the best average performance in all cases.


\begin{table}[t]
\centering
\scriptsize
\setlength\tabcolsep{2pt}
\caption{Attack success rate (\%) of RN50-generated adversarial examples under adversarial defenses. The bold item is the best one. The AVG column indicates the average attack success rate under defense.}
\label{tab:defense}
\begin{tabular}{ccccccccc} 
\hline  
        	  & AdvIncV3      & AdvIncResV2$_{ens}$    & JPEG          & NRP           & RS            & NIPS-r3        & R\&P          & AVG               \\ \hline
MIFGSM        & 43.9          & 25.0            & 95.7          & 96.5          & 25.4          & 32.7          & 79.2          & 56.91             \\
DIFGSM        & 35.5          & 20.9          & 95.6          & \textbf{98.8} & 22.7          & 27.3          & 74.0            & 53.54             \\
TIFGSM        & 36.5          & \textbf{31.6} & 95.9          & 89.2          & \textbf{26.9} & 37.1          & 77.9          & 56.44             \\
NIFGSM        & 44.4          & 25.4          & \textbf{96.8} & 92.3          & 26.7          & 32.0            & 78.1          & 56.53         	  \\
SI$^2$FGSM    & 40.6          & 28.7          & 96.1          & 97.6          & 22.2          & 31.6          & 77.7          & 56.36         	 \\
FAUG          & \textbf{48.0}   & 29.4          & 95.6          & 95.2          & 26.1          & \textbf{38.4} & \textbf{80.2} & \textbf{58.99}	 \\  \midrule\hline\noalign{\smallskip} 
VNIFGSM       & 52.4          & 36.3          & \textbf{96.8} & \textbf{97.5} & \textbf{27.6} & 44.6          & 83.6          & 62.69         	 \\
VNIFGSM-FAUG  & \textbf{55.4} & \textbf{41.1} & 96.6          & 95.7          & 27.0            & \textbf{48.6} & \textbf{83.9} & \textbf{64.04}	 \\ \hline
VMIFGSM       & 49.9          & 35.2          & \textbf{96.2} & \textbf{97.6} & 26.3          & 45.5          & 81.7          & 61.77         	 \\
VMIFGSM-FAUG  & \textbf{53.0}   & \textbf{39.5} & 96.0            & 96.3          & \textbf{27.7} & \textbf{48.4} & \textbf{84.0}   & \textbf{63.56}	 \\ \hline
SINIFGSM      & 51.4          & 36.2          & 97.4          & 82.6          & 28.5          & 44.4          & 84.7          & 60.74         	 \\
SINIFGSM-FAUG & \textbf{54.5} & \textbf{40.9} & \textbf{97.5} & \textbf{82.8} & \textbf{30.8} & \textbf{51.3} & \textbf{86.3} & \textbf{63.44}	 \\ \hline
AMVIFGSM      & 62.7          & \textbf{51.4} & 96.6          & \textbf{88.0}   & 34.5          & 59.7          & 87.4          & 68.61         	 \\
AMVIFGSM-FAUG & \textbf{63.8} & 50.4          & \textbf{97.4} & 87.0            & \textbf{35.6} & \textbf{61.0}   & \textbf{87.8} & \textbf{69.00}	 \\ \hline
\end{tabular}
\end{table}

\begin{table}[t]
\centering
\tiny
\setlength\tabcolsep{2pt}
\caption{Comparison results of standard ensemble attacks and the FAUG enhanced counterpart in terms of attack success rate (\%). The bold item indicates the best one; the item with $\star$ superscript is the white-box attack, and the others are black-box attacks. CNN denotes ensemble four convolution models, and Transformer denotes ensemble four transformer models. AVG column indicates the average attack success rate on black-box models.}
\label{tab:ensemble}
\begin{tabular}{c|c|ccccccccc}
\hline
Ensemble Models                 &                & RN50 & DN121 & VGG19BN & RNX50 & Visformer-S & PiT-B & ViT-B/16 & Swin-B/S3 & AVG   \\ \hline
\multirow{2}{*}{CNN}                            & Standard & 97.5$^\star$ & 96.7$^\star$  & 96.6$^\star$    & 98.2$^\star$  & 82.8        & 64.2  & 45.7     & 56.3      & 62.25 \\
                                                & FAUG   & \textbf{98.0}$^\star$   & \textbf{97.0}$^\star$    & 96.6$^\star$    & \textbf{98.5}$^\star$  & \textbf{88.4}        & \textbf{76.1} & \textbf{59.9} & \textbf{71.1} & \textbf{73.88} \\ \hline
\multirow{2}{*}{Transformer}                     & Standard & 76.2 & 77.3  & 78.2    & 78.0    & 96.7$^\star$    & \textbf{96.3}$^\star$  & 93.9$^\star$  & \textbf{97.2}$^\star$      & 77.43 \\
                                                & FAUG   & \textbf{85.8} & \textbf{87.5} & \textbf{86.9} & \textbf{87.6} & \textbf{97.6}$^\star$ & 95.4$^\star$  & \textbf{94.2}$^\star$ & 93.8$^\star$  & \textbf{86.95} \\ \hline
\multirow{2}{*}{\shortstack{RN50+VGG19BN\\PiT-B+Visformer-S}} & Standard & \textbf{97.6}$^\star$ & 93.4  & 96.6$^\star$    & 94.9  & 98.0$^\star$          & 96.2$^\star$  & 52.4     & 77.2      & 79.48 \\
                                                & FAUG  & 97.4$^\star$  & \textbf{94.8} & \textbf{96.7}$^\star$    & \textbf{96.2} & \textbf{98.5}$^\star$   & 96.2$^\star$   & \textbf{63.1} & \textbf{86.9} & \textbf{85.25} \\ \hline
\end{tabular}
\end{table}

\subsection{Evaluation on Ensemble Model}
The previous works \cite{mim2018BoostingAA,tramer2018ensemble} have demonstrated that attacking multiple models simultaneously (i.e., ensemble attack) can significantly boost adversarial transferability. Thus, it is necessary to study whether the proposed FAUG can further improve the adversarial transferability in ensemble attacks. Specifically, we investigate three ensemble strategies: four CNN model ensembles (abbr. CNN), four transformer model ensembles (abbr. Transformer), four model ensembles composed by two CNN (RN50 and VGG19BN) and Transformer (PiT-B and Visformer-S). We perform an average fusion of four models on the logit layer (i.e., the layer before the softmax). Table \ref{tab:ensemble} reports the comparison results. As we can observe, ensemble attacks enhanced with the proposed FAUG consistently surpass the standard ensemble attacks, gaining improvement of 11.63\%, 9.53\%, and 5.77\% on three ensemble strategies in terms of average transferability rate. Such a result suggests the effectiveness of the proposed feature augmentation method. On the other side, the four transformer models' ensemble attacks exhibit more efficiency than the other two ensemble strategies in improving the adversarial transferability. Concretely, the Transformer achieves the highest adversarial transferability of 86.95\% on black-box models, while the CNN is 73.88\%.

\begin{figure}[t]
	\centering
	\begin{minipage}{.45\linewidth}
		\centering
		\includegraphics[width =1.\linewidth]{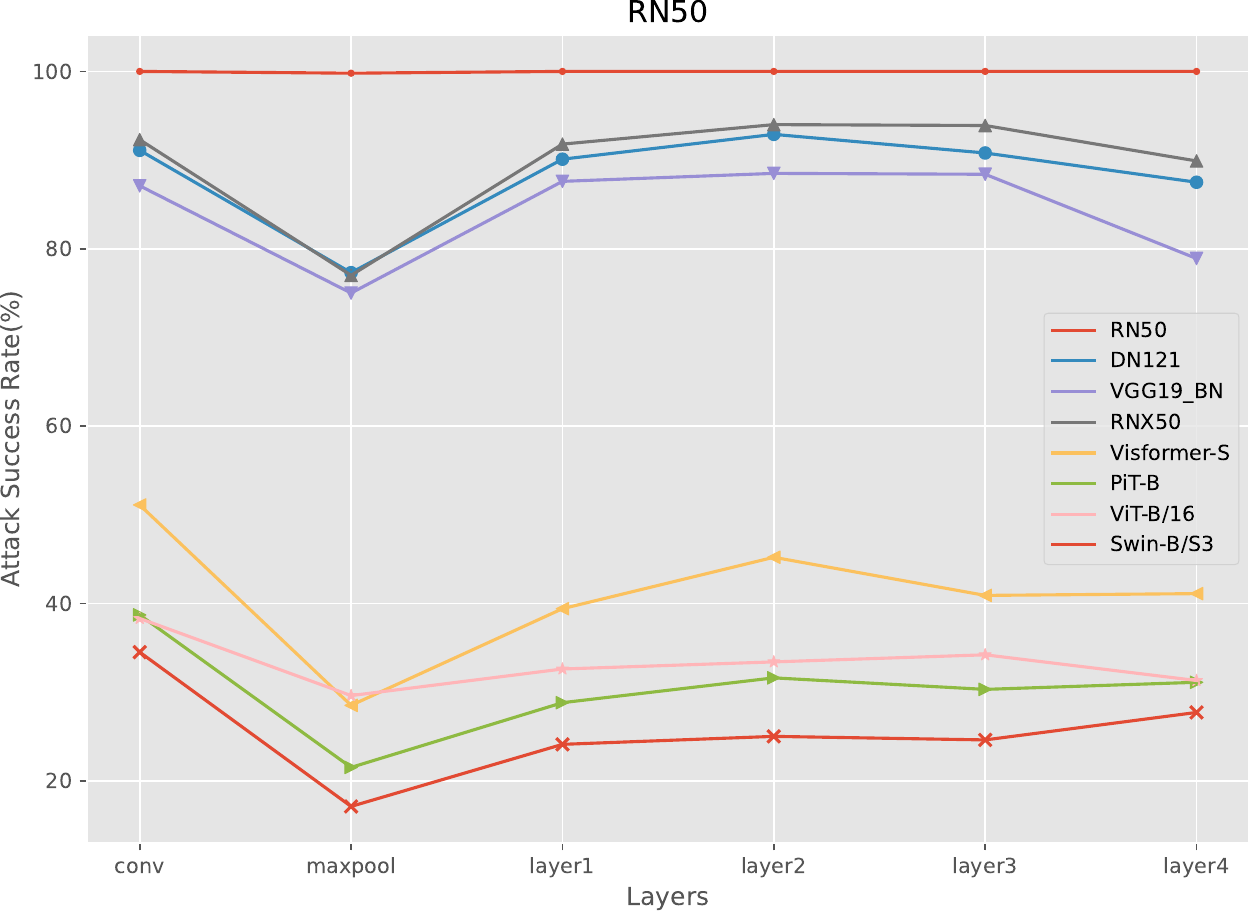}
	\end{minipage}
	\begin{minipage}{.45\linewidth}
		\centering
		\includegraphics[width =1.\linewidth]{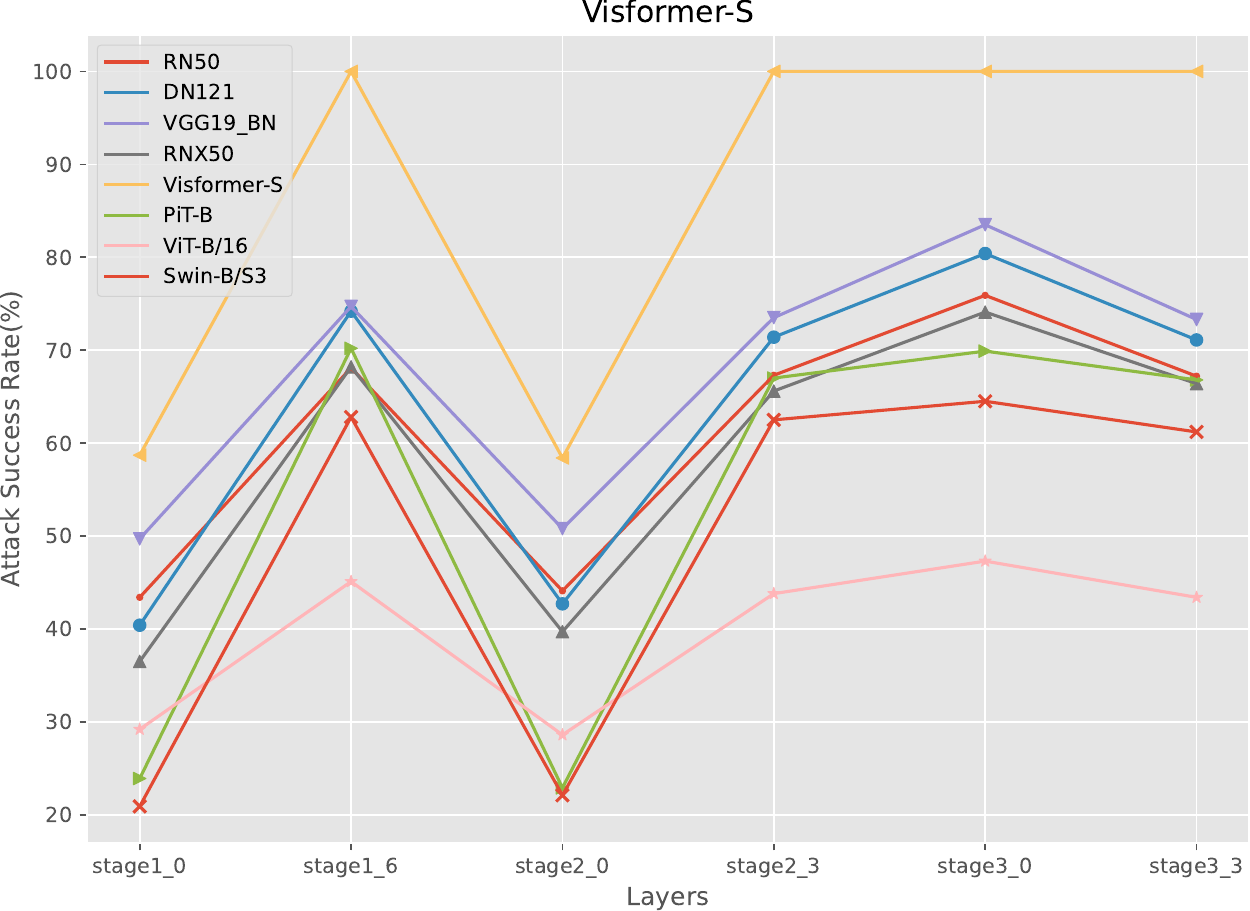}
	\end{minipage}
\caption{Influence of layer selection on attack performance.}
\label{fig:abla_layer}
\end{figure}

\subsection{Ablation Study}
\label{sec:abla}
To comprehensively investigate potential factors that may influence our method, we conduct three ablation studies in this section: the influence of layer selection, feature augmentation type, and noise strength ($\sigma$) of random normal noise. 

{\bfseries Influence of the layer selection.}
To investigate the influence of layer selection on adversarial transferability, we conducted experiments using six layers from RN50 and Visformer-S, and set the standard deviation $\sigma$ to 3.0. Then, we introduce the random noise into different intermediate layers and generate corresponding adversarial examples. Figure \ref{fig:abla_layer} illustrates the evaluation results. As we can observe, on the one hand, RN50 and Visformer-S exhibit significant discrepancies, such as 1) RN50 shows a slight fluctuation across different layers and achieves the best average transferability rate of 61.87\% at the \textsf{conv} layer. This outcome may be due to the preservation of rich texture or edge information in the shallow layer of RN50, which are retained in various feature channels as stated in \cite{wang2023improving}. Thus,  performing feature augmentation in the shallow layer resembles input augmentation but is more efficient since it augments all channels. In contrast, 2) Visformer-S shows a large fluctuation, markedly different from RN50, which may be attributed to the input processing formal of the transformer, where the image is chunked into multiple image patches. Nonetheless, the \textsf{stage3\_0} layer of Visformer-S archives the highest average transferability rate of 74.24\%, surpassing RN50. On the other hand, the layer selection significantly impacts adversarial transferability within specific models. For instance, the follow-up intermediate layers of RN50 show an increasing trend followed by a decline. The maximum gap in average transferability rate between the best (i.e., 61.87\%) and the worst (i.e., 46.57\%  engendered by \textsf{maxpool}) is 15.3\%. Similarity, the counterpart of Visformer-S is 37.2\%, and the worst and best is produced by \textsf{stage1\_0} (i.e., 37.04\%) and \textsf{stage3\_0} layer (i.e., 74.24\%), respectively. Therefore, we select the \textsf{conv} layer and \textsf{stage3\_0} layer for RN50 and Visformer-S, respectively. Further details regarding the layer names and selection for other models are provided in Supplementary Sec. \ref{supp:layer}.

\begin{figure}[t]
	\centering
	\begin{minipage}{.45\linewidth}
		\centering
		\includegraphics[width =1.\linewidth]{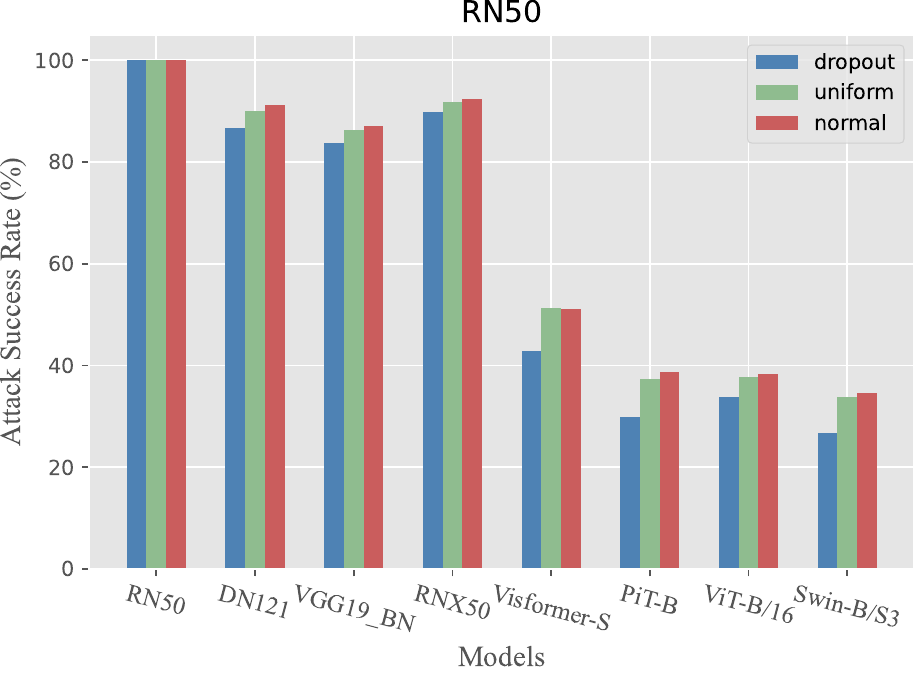}
	\end{minipage}
	\begin{minipage}{.45\linewidth}
		\centering
		\includegraphics[width =1.\linewidth]{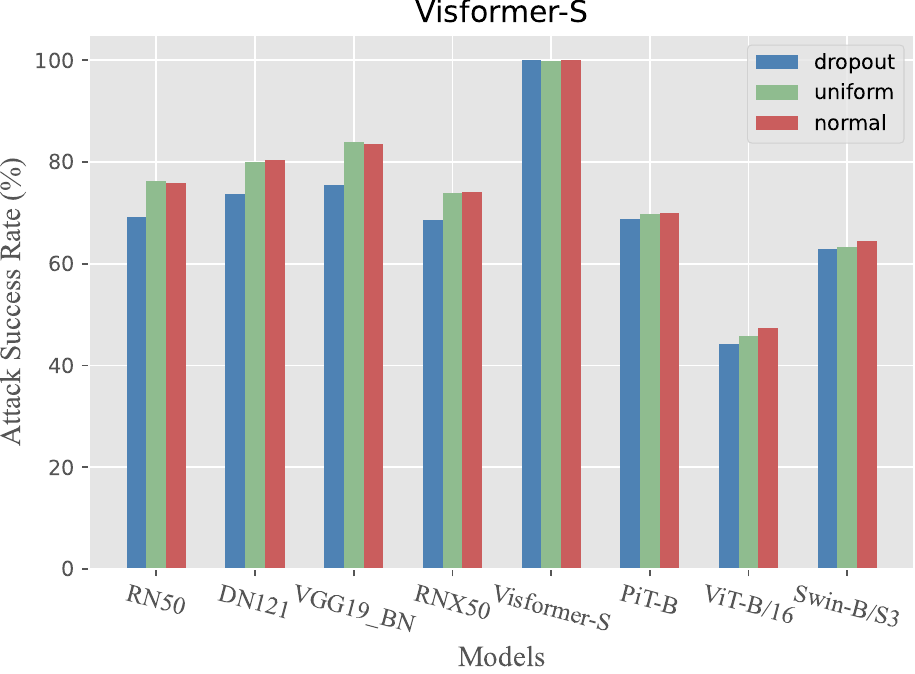}
	\end{minipage}
\caption{Influence of the type of feature augmentation on attack performance.}
\label{fig:abla_aug_type}
\end{figure}

{\bfseries Influence of feature augmentation type.}
To investigate the influence of different feature augmentation types on adversarial transferability, we replaced the random normal noise with two alternative augmentation types: random uniform noise and dropout operator. Specifically, we conducted experiments on RN50 and Visformer-S, and set the lower/upper bound of uniform noise to $\pm0.3$ and $\pm0.2$, the probability of the dropout operator is set to 0.3. Figure \ref{fig:abla_aug_type} illustrates the evaluation results. As one can see, random normal noise achieves superior adversarial transferability on both RN50 and Visformer-S compared to the other two methods. For RN50, the average transferability rates are 61.87\%, 61.19\%, and 56.21\% by using normal, uniform, and dropout feature augmentation. Similarly, for Visformer-S, the corresponding rates are 70.8\%, 70.39\%, and 66.09\%. The differences in transferability between different types of random noises are relatively small, such as the discrepancy between normal and uniform noise is 0.68\% and 0.41\% at similar noise strengths, respectively, suggesting that the proposed feature augmentation method is not limited to specific random noise types. Based on these findings, we opt to use random normal noise in our feature augmentation attack.

\begin{figure}[t]
	\centering
	\begin{minipage}{.45\linewidth}
		\centering
		\includegraphics[width =1.\linewidth]{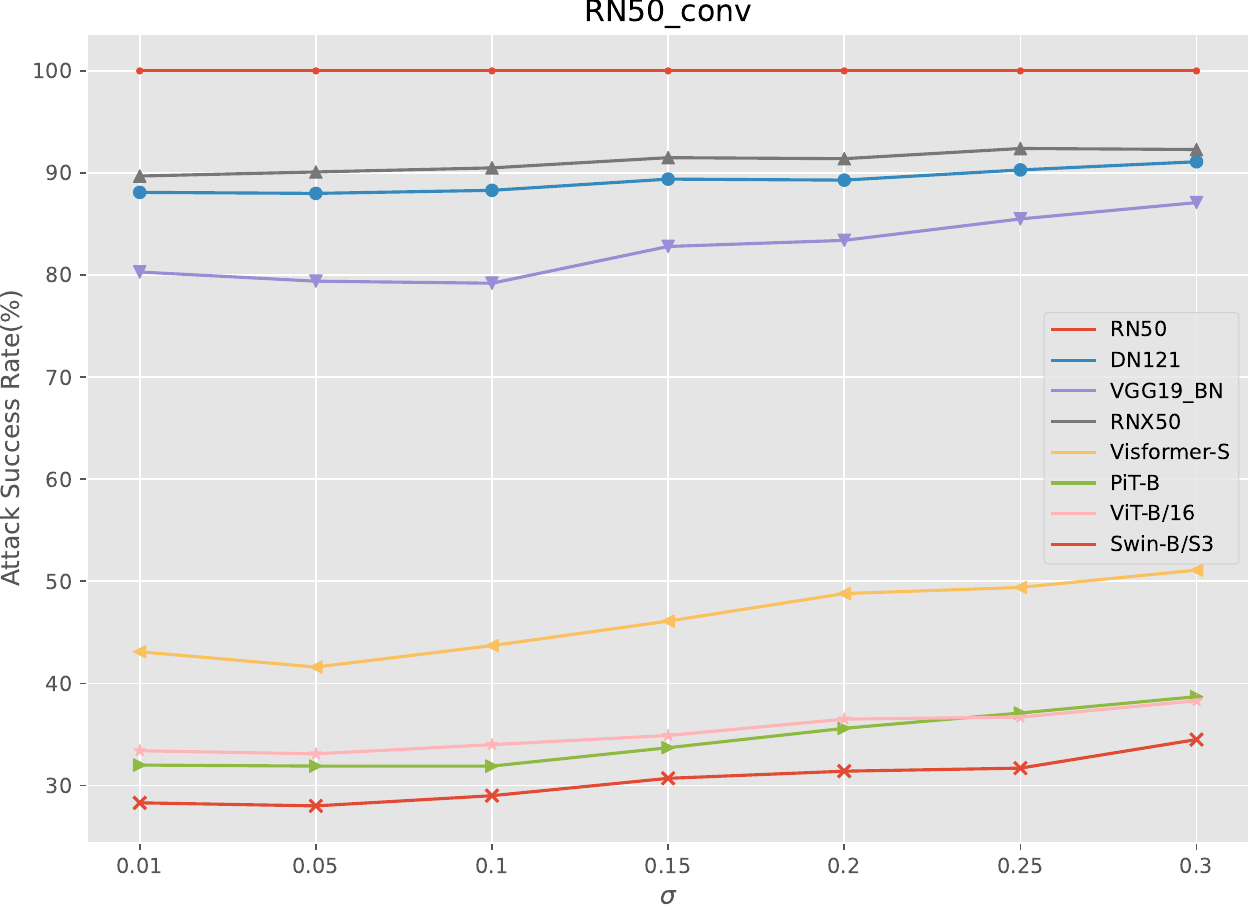}
	\end{minipage}
	\begin{minipage}{.45\linewidth}
		\centering
		\includegraphics[width =1.\linewidth]{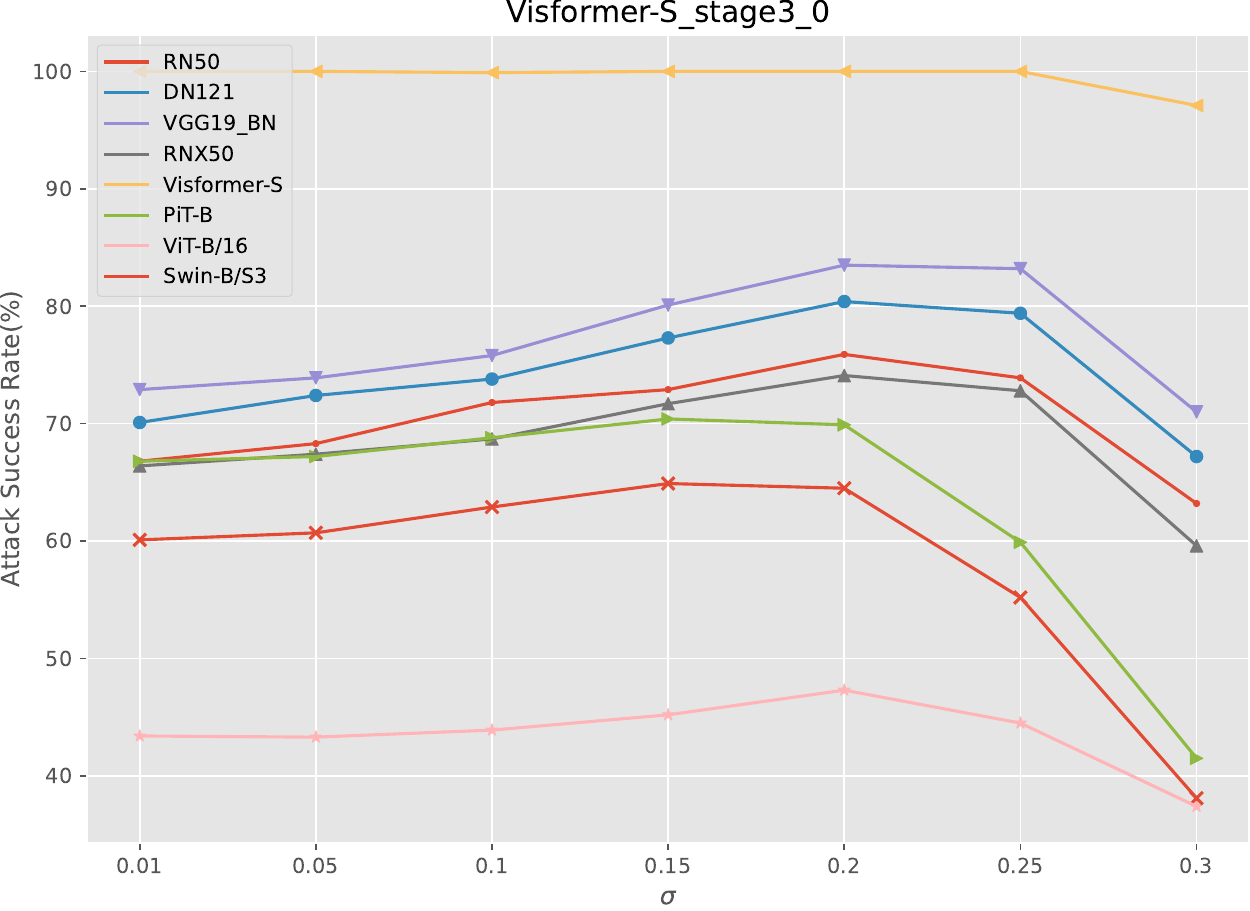}
	\end{minipage}
\caption{Influence of random noise strength on attack performance at the specific layer.}
\label{fig:abla_std}
\end{figure}

{\bfseries Influence of random noise strength.}
Feature augmentation under different noise strengths has varying degrees of influence on feature discrepancy and leads to varying adversarial transferability. To reveal such a relationship, we conduct a grid search on standard deviation $\sigma$ of the random noise within the range of $\left\{0.01, 0.05, 0.1, 0.15, 0.2, 0.25, 0.3\right\}$. This was performed on \textsf{conv} and \textsf{stage3\_0} layer on RN50 and Visformer-S, respectively. Figure \ref{fig:abla_std} depicts the evaluation results. As we can see, RN50 displays a relatively stable increasing trend in average transferability rate with increasing noise strength, reaching a peak of 61.87\% at $\sigma$ is 0.3. In contrast, Visformer-S exhibits an initial increase followed by a decrease in transferability rate, peaking at 74.24\% when $\sigma=0.2$. This suggests that the optimal noise strength varies between models, indicating the importance of tuning this parameter for maximizing adversarial transferability. Consequently, for RN50 and Visformer-S, we set the noise strength at \textsf{conv} and \textsf{stage3\_0} layer to 0.3 and 0.2, respectively. Further details on other models can be found in Supplementary Sec. \ref{supp:std}.

\section{Social Impact and Limitation}
{\bfseries Social Impact} Our work endeavors to improve adversarial transferability, which may impose potential security risks for those DNN-based applications. However, our method can be countered by adopting more sophisticated defense mechanisms, e.g., noise elimination, noise detection, or denoiser method. Furthermore, our generated adversarial examples can serve as a novel training set for adversarial training, enhancing adversarial robustness.

{\bfseries Limitation} As evident from the ablation study, our feature augmentation attack exhibits sensitivity to both layer selection and noise strength, which vary depending on the specific attack and model. Consequently, the careful selection of layers and optimal noise strength is crucial for achieving improved adversarial transferability.


\section{Conclusion}
\label{sec:conclusion}
In this paper, we propose a feature augmentation method designed to substantially enhance adversarial transferability. Specifically, we inject random noise into the intermediate feature of the model to amplify the diversity of the attack gradient, mitigating overfitting and leading to better transferability. Our feature augmentation method can be seamlessly integrated with advanced gradient and ensemble attacks to further enhance the attack performance. Note that our method incurs no additional computational resources apart from adding random noise to the feature. Extensive experiments conducted on ImageNet across CNN and transformer models demonstrate the effectiveness of the proposed method. 


\par\vfill\par
\clearpage  

%
%
\bibliographystyle{splncs04}
\bibliography{main}

\end{document}


\title{Supplementary: Improving the Transferability of Adversarial Examples by Feature Augmentation} 

\author{Donghua Wang\inst{1,2} \and
Wen Yao\inst{2,4} \and
Tingsong Jiang \inst{2,4} \and
Xiaohu Zheng \inst{2,4} \and
Junqi Wu \inst{3,4} \and
Xiaoqian Chen \inst{2,4}
}

\authorrunning{Wang et al.}

\institute{Zhejiang University \and
Defense Innovation Institute, Chinese Academy of Military Science \and
Shanghai Jiao Tong Unversity \and
Intelligent Game and Decision Laboratory
}

\maketitle

\section{Implementation details}
\label{supp:details}
In this section, we presents the implementation details of the proposed feature augmentation, which includes the model name, the corresponding abbreviation name, the pre-trained checkpoint file provided by \textsf{timm} package, the intermediate layer which random normal noise is added, and the standard deviation($\sigma$). Here, we elucidate the meaning of the name of the intermediate layer. For RN50 and RNX50, \textsf{conv} refers to the first convolutional layer of the model. For DN121 and VGG19BN, \textsf{feature\_0} denotes the first layer of the feature extraction module. For Visformer-S, \textsf{stage3\_0} indicates the 0-th block of the 3-th transformer module. For PiT-B, \textsf{block\_1\_0} indicates the 0-th block of the 1-th transformer module. For ViT-B/16, \textsf{stage\_5} indicates the 5-th transformer module. For Swin-B/S3, \textsf{stage\_0\_0} indicates the 0-th block of the 0-th transformer module.

\begin{table}[]
\centering
\tiny
\setlength\tabcolsep{2pt}
\caption{Implementation details of the feature augmentation.}
\label{supp:tab_param}
\begin{tabular}{c|cccc}
\hline
Model                   & Abbreviation & Pretrained Checkpoint                & Perturbed Layer & $\sigma$ \\ \hline
resnet50                & RN50         & resnet50-19c8e357.pth                & conv            & 0.3   \\
densenet121             & DN121        & densenet121-a639ec97.pth             & feature\_0      & 0.3   \\
vgg19\_bn               & VGG19BN      & vgg19\_bn-c79401a0.pth               & feature\_0      & 0.25  \\
resnext50\_32x4d        & RNX50        & resnext50\_32x4d-7cdf4587.pth        & conv            & 0.3   \\
visformer\_small        & Visformer-S  & timm/visformer\_small\_in1k.bin      & stage3\_0       & 0.2   \\
pit\_b\_224             & PiT-B        & timm/pit\_b\_224\_in1k.bin           & block\_1\_0     & 0.3   \\
vit\_base\_patch16\_224 & ViT-B/16     & jx\_vit\_base\_p16\_224-80ecf9dd.pth & block\_5        & 0.8   \\
swin\_s3\_base\_224     & Swin-B/S3    & s3\_b-a1e95db4                       & stage\_0\_0     & 0.6  \\ \hline
\end{tabular}
\end{table}

\clearpage
Here, we provide the following pseudocode to describe the implementation of our method in performing the feature augmentation on the first \textsf{conv} layer of RN50.
\begin{python}
import torch
import torch.nn as nn

# Feature Augmented Model
class ResNet50(nn.Module):
	def __init__(self, *args, **kwargs):
		self.conv = ...
		self.maxpool = ...
		...
		self.layer1 = ...
		self.layer2 = ...
		self.layer3 = ...
		self.layer4 = ...
		...
		
	def forward(self, x):
		x = self.conv(x)
		# add the random noise into the first conv layer
		x.data += torch.zeros_like(x).normal_(mean=0, std=0.3)
		...
		
		return x
		
\end{python}

\section{Influence of the layer choice}
\label{supp:layer}

\begin{figure}[t]
	\centering
	\centering
	\begin{minipage}{.22\linewidth}
		\centering
		\includegraphics[width =1.\linewidth]{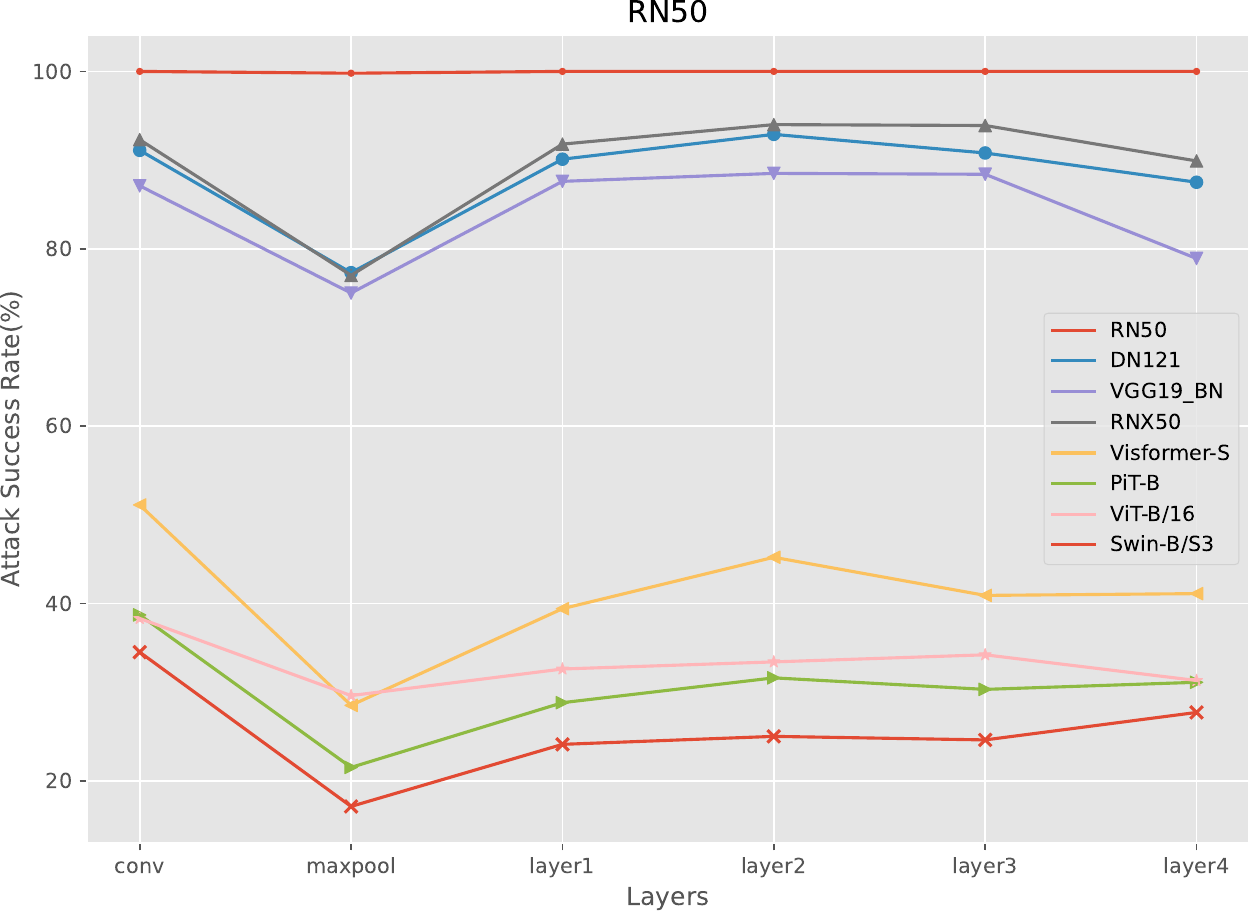}
	\end{minipage}
	\begin{minipage}{.22\linewidth}
		\centering
		\includegraphics[width =1.\linewidth]{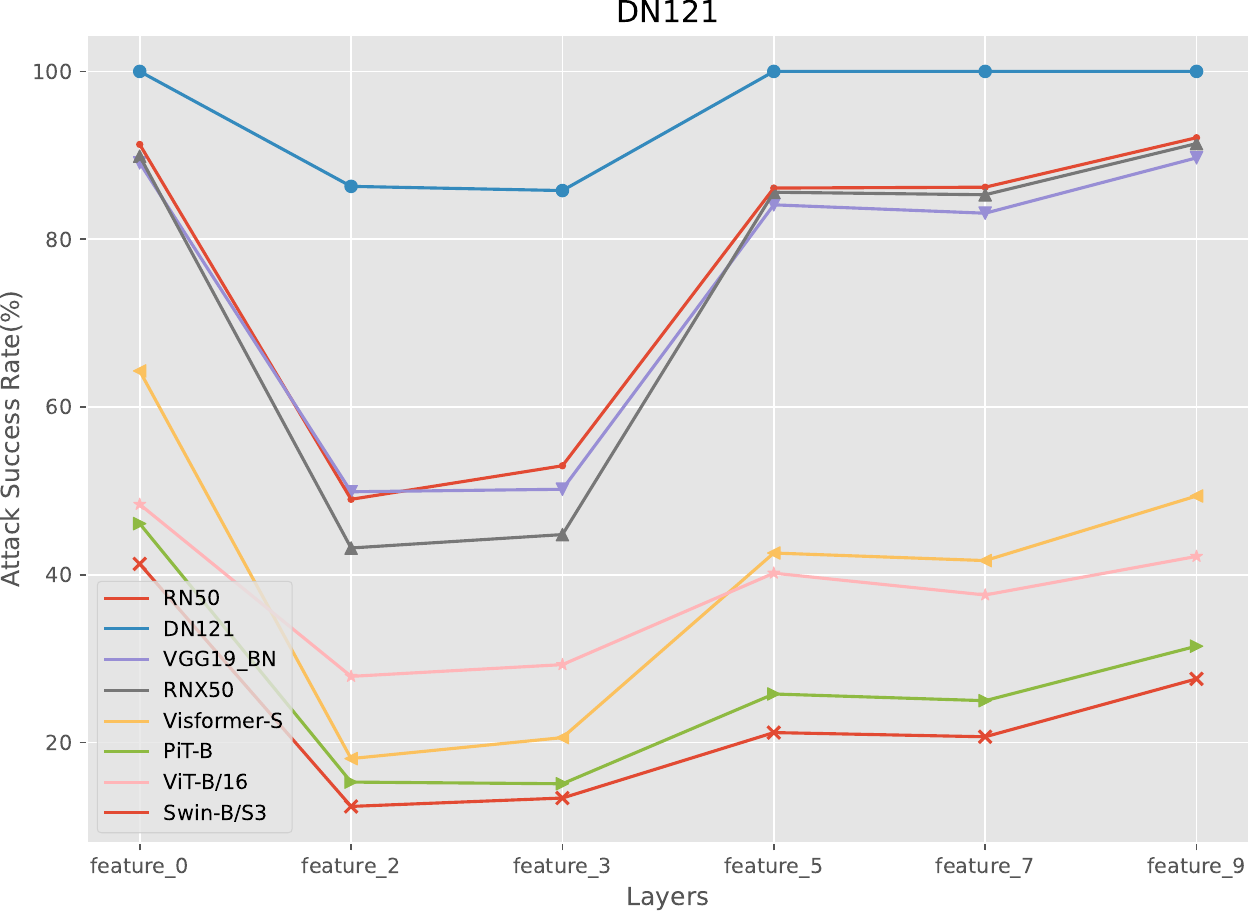}
	\end{minipage}
	\begin{minipage}{.22\linewidth}
		\centering
		\includegraphics[width =1.\linewidth]{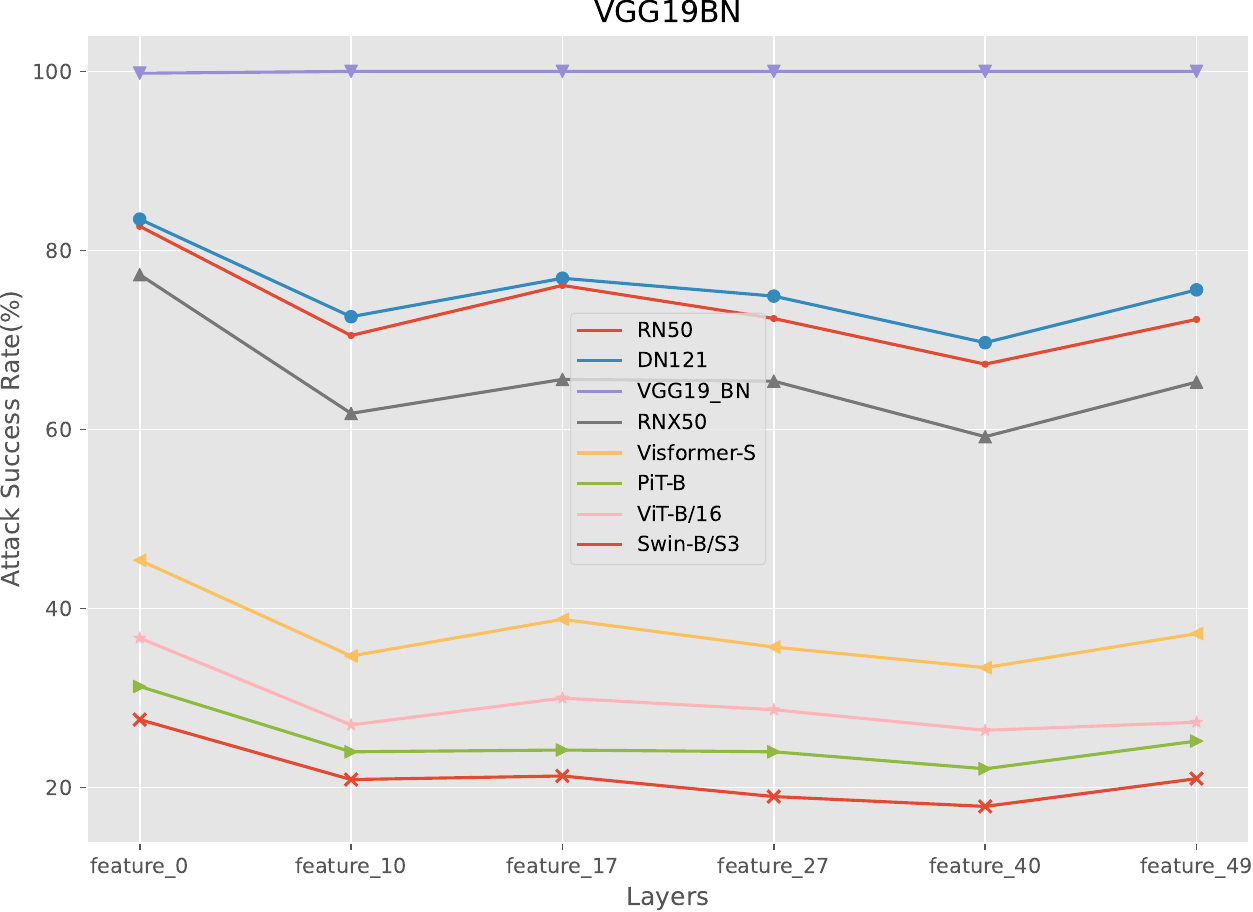}
	\end{minipage}
	\begin{minipage}{.22\linewidth}
		\centering
		\includegraphics[width =1.\linewidth]{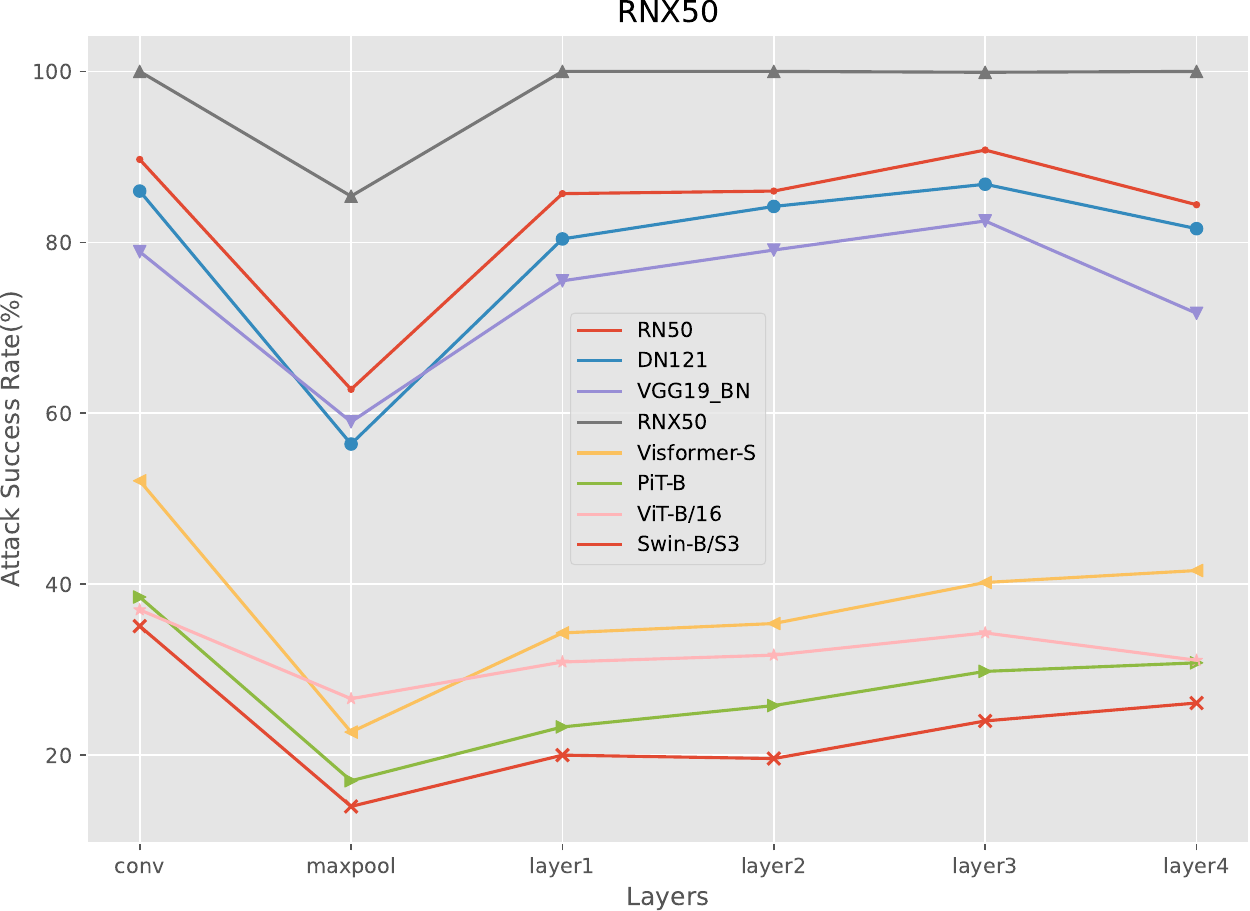}
	\end{minipage}
	
	\begin{minipage}{.22\linewidth}
		\centering
		\includegraphics[width =1.\linewidth]{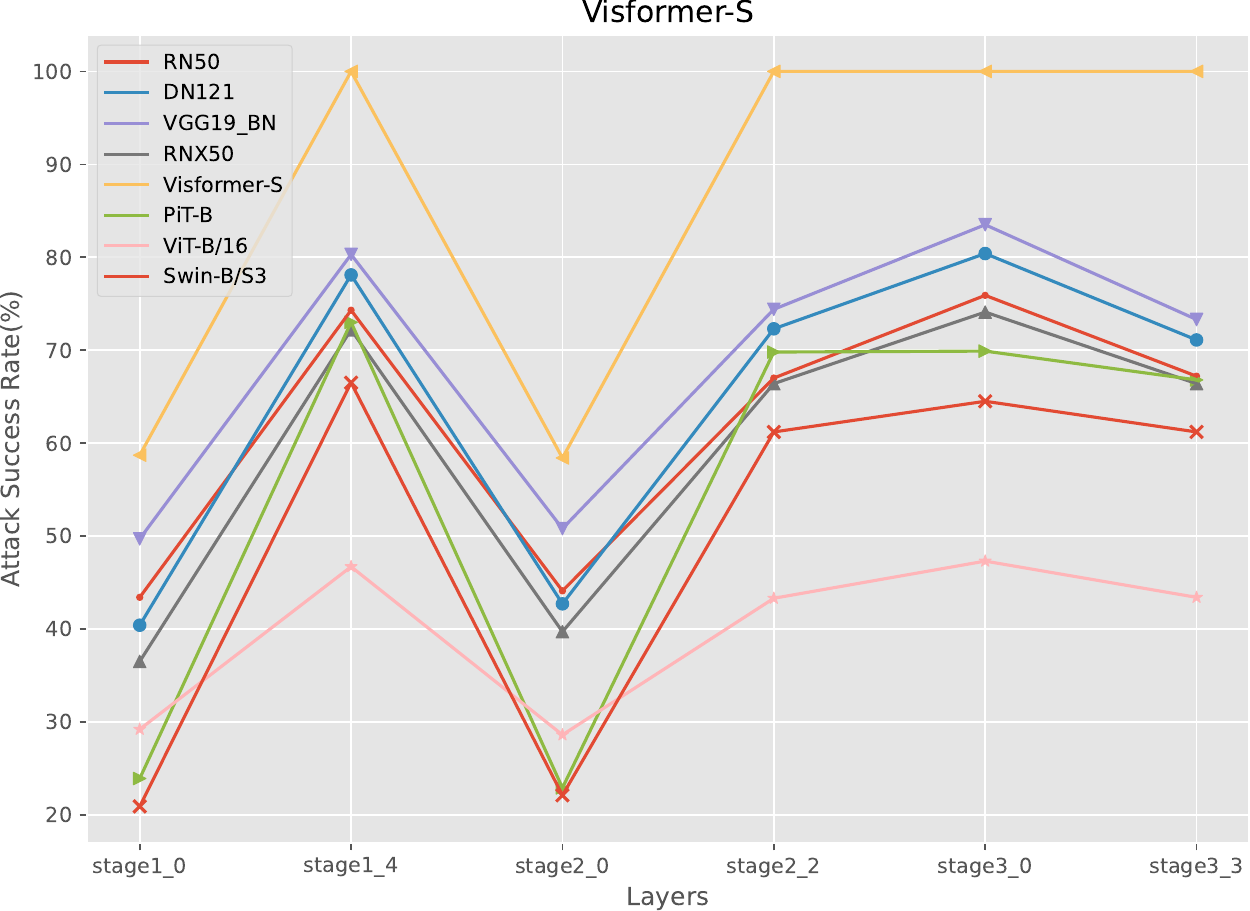}
	\end{minipage}
	\begin{minipage}{.22\linewidth}
		\centering
		\includegraphics[width =1.\linewidth]{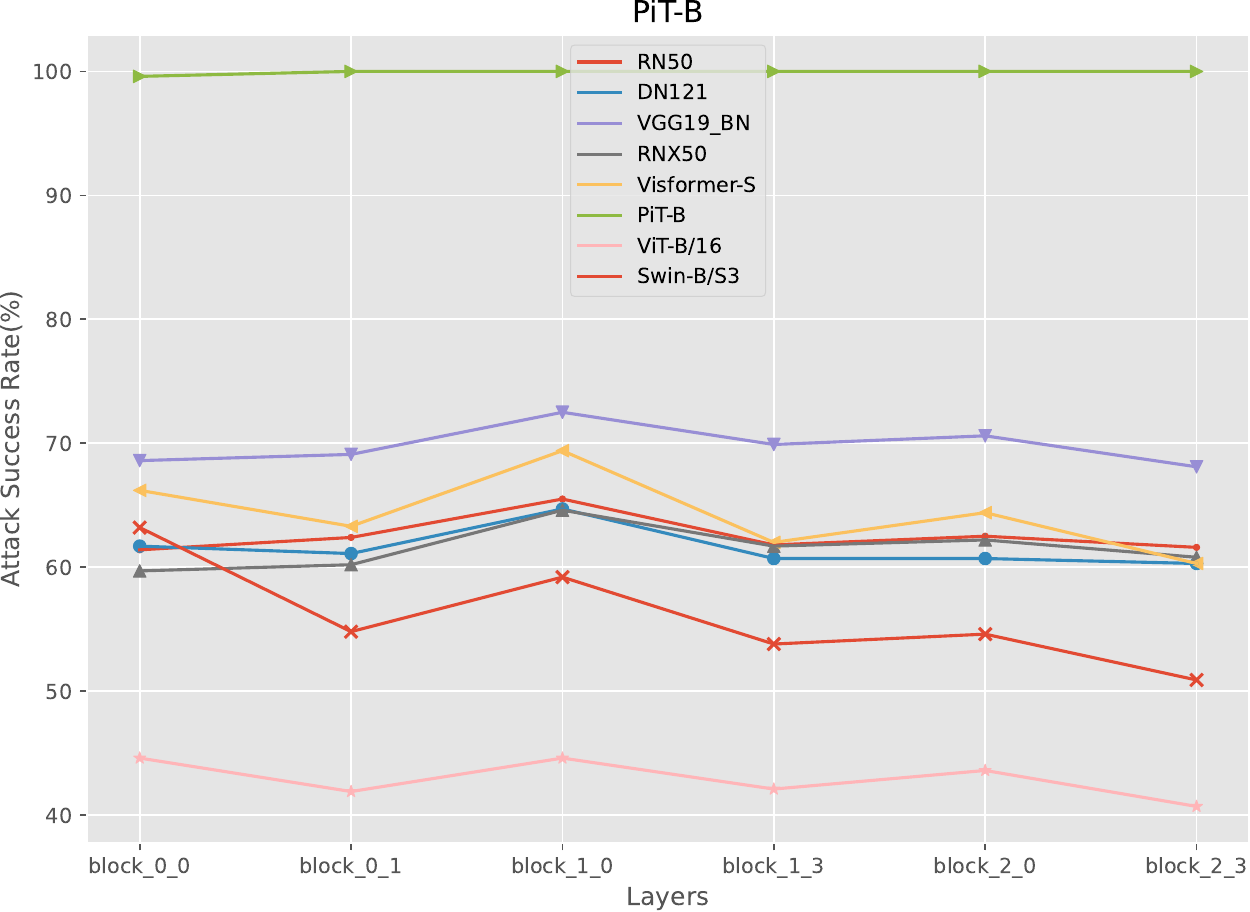}
	\end{minipage}
	\begin{minipage}{.22\linewidth}
		\centering
		\includegraphics[width =1.\linewidth]{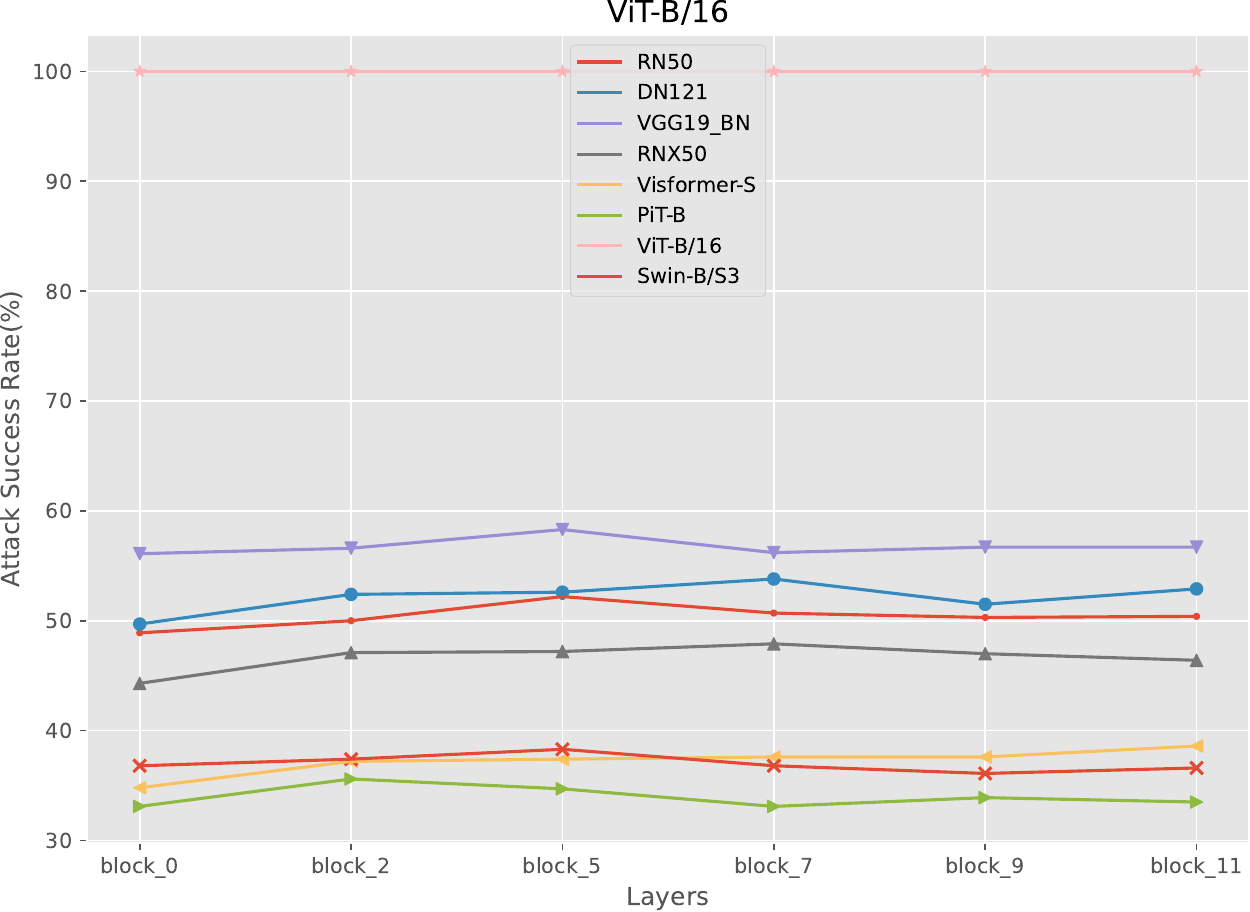}
	\end{minipage}
	\begin{minipage}{.22\linewidth}
		\centering
		\includegraphics[width =1.\linewidth]{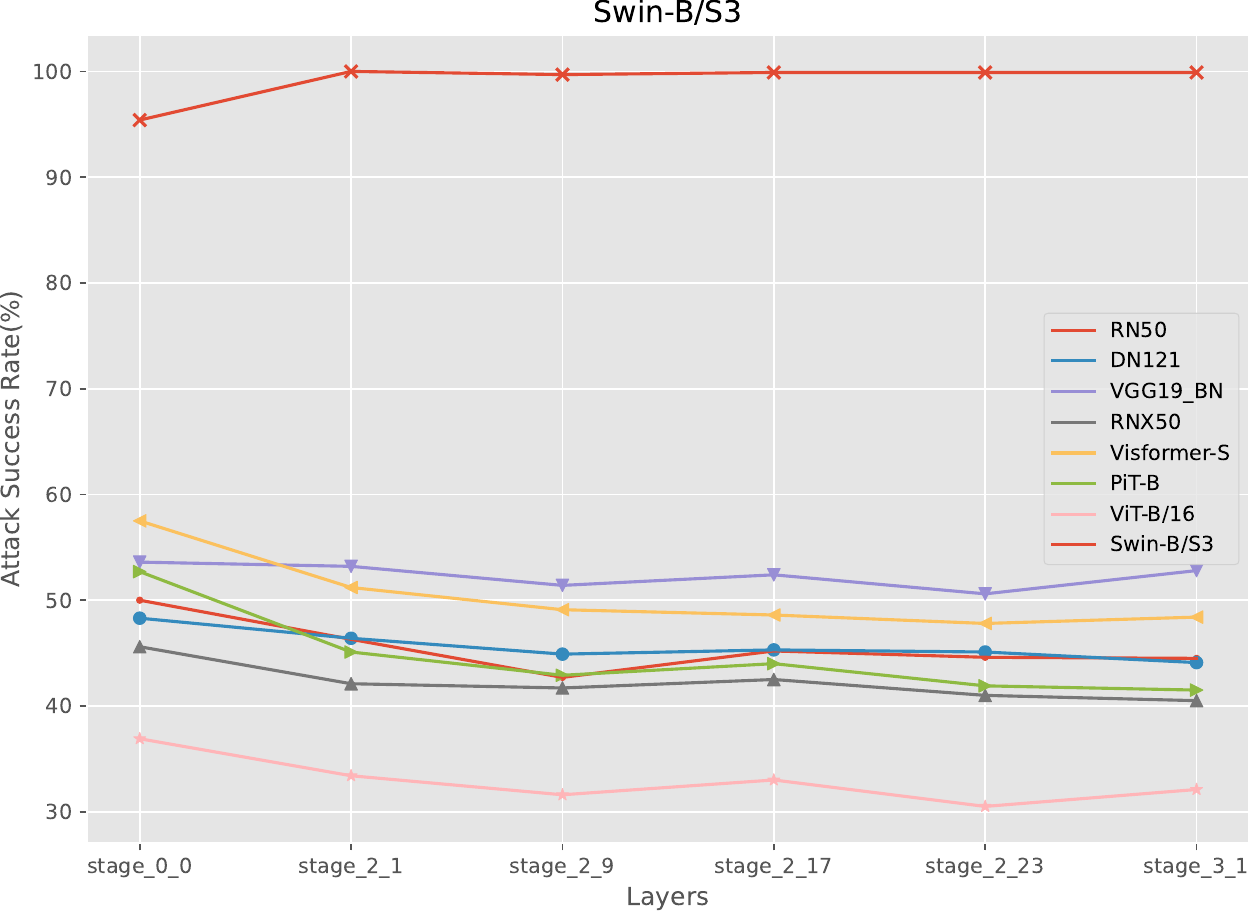}
	\end{minipage}
\caption{Influence of layer choice on attack performance.}
\label{fig:supp_abla_layer}
\end{figure}

In this part, we provide detailed ablation results on the influence of layer selection on adversarial transferability across eight models, We conduct the experiments by fixing the $\sigma$ value as reported in Table \ref{supp:tab_param}. The ablation results are illustrated in Figure \ref{fig:supp_abla_layer}. As we can see, the first layer of the CNN model leads to a better average transferability rate, while the Transformer is specificity. Specifically, the best average transferability rates of four CNN models (RN50, DN121, VGG19BN, RNX50) are obtained at the first layer, with values of 61.87\%, 68.44\%, 55.4\%, and 59.19\%, respectively. In contrast, the corresponding values for the four Transformer models (e.g., Visformer-S, PiT-B, ViT-B/16 and Swin-B/S3) are 71.06\%, 63\%, 47.94\% and 54.16\%, achieved at \textsf{stage3\_0}, \textsf{block\_1\_0}, \textsf{block\_5}, \textsf{stage\_0\_0}, correspondingly. Moreover, we can also observe that different models exhibit varying trends; for example, Visformer-S shows fierce fluctuation, and ViT-B/16 shows a stable trend. In summary, the ablation study shows that seven out of eight models are prone to yielding the best results in the shallow layer, which supports our claim that the feature augmentation on the shallow layer is assumably equivalent to the input augmentation but in a more efficient manner.

\section{Influence of random noise strength}
\label{supp:std}

\begin{figure}[t]
	\centering
	\begin{minipage}{.22\linewidth}
		\centering
		\includegraphics[width =1.\linewidth]{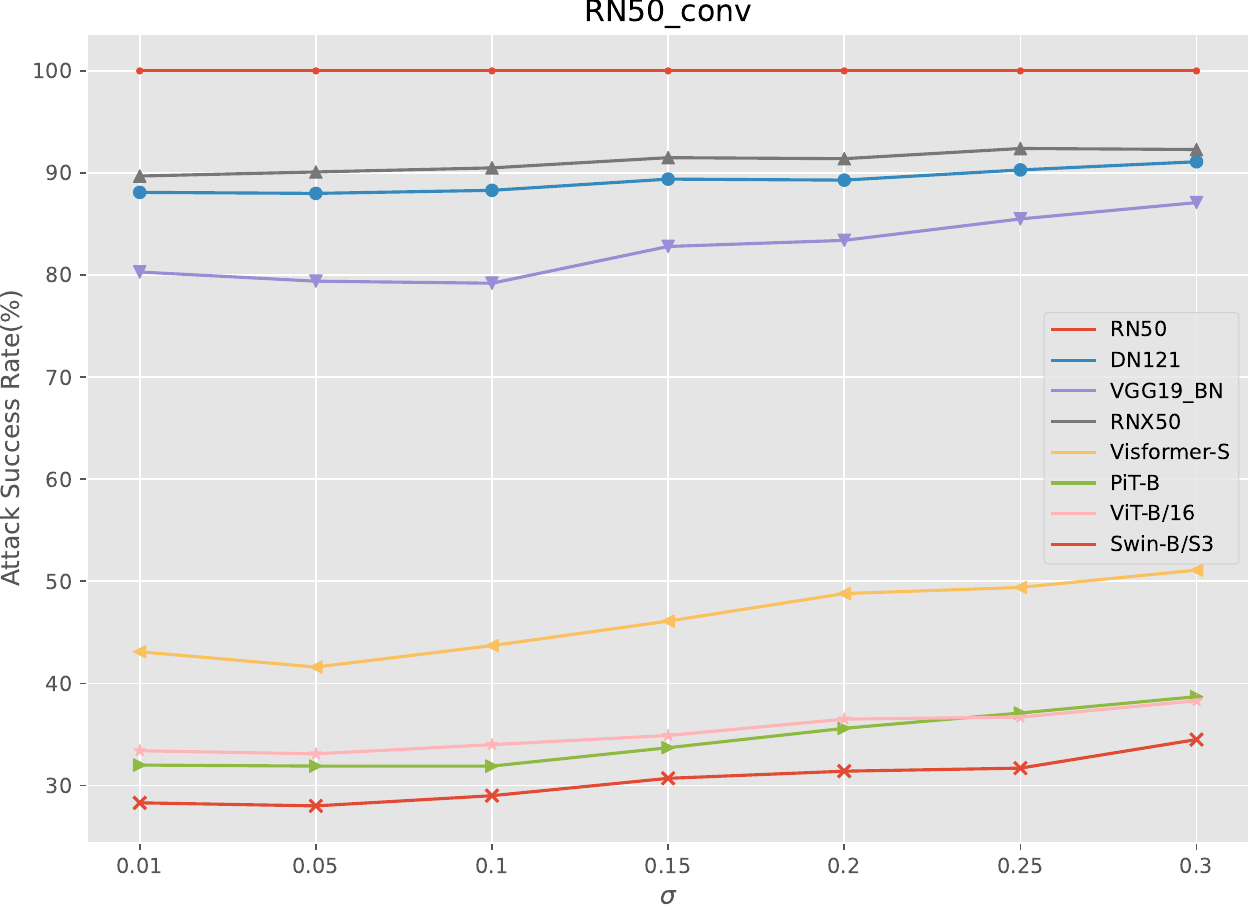}
	\end{minipage}
	\begin{minipage}{.22\linewidth}
		\centering
		\includegraphics[width =1.\linewidth]{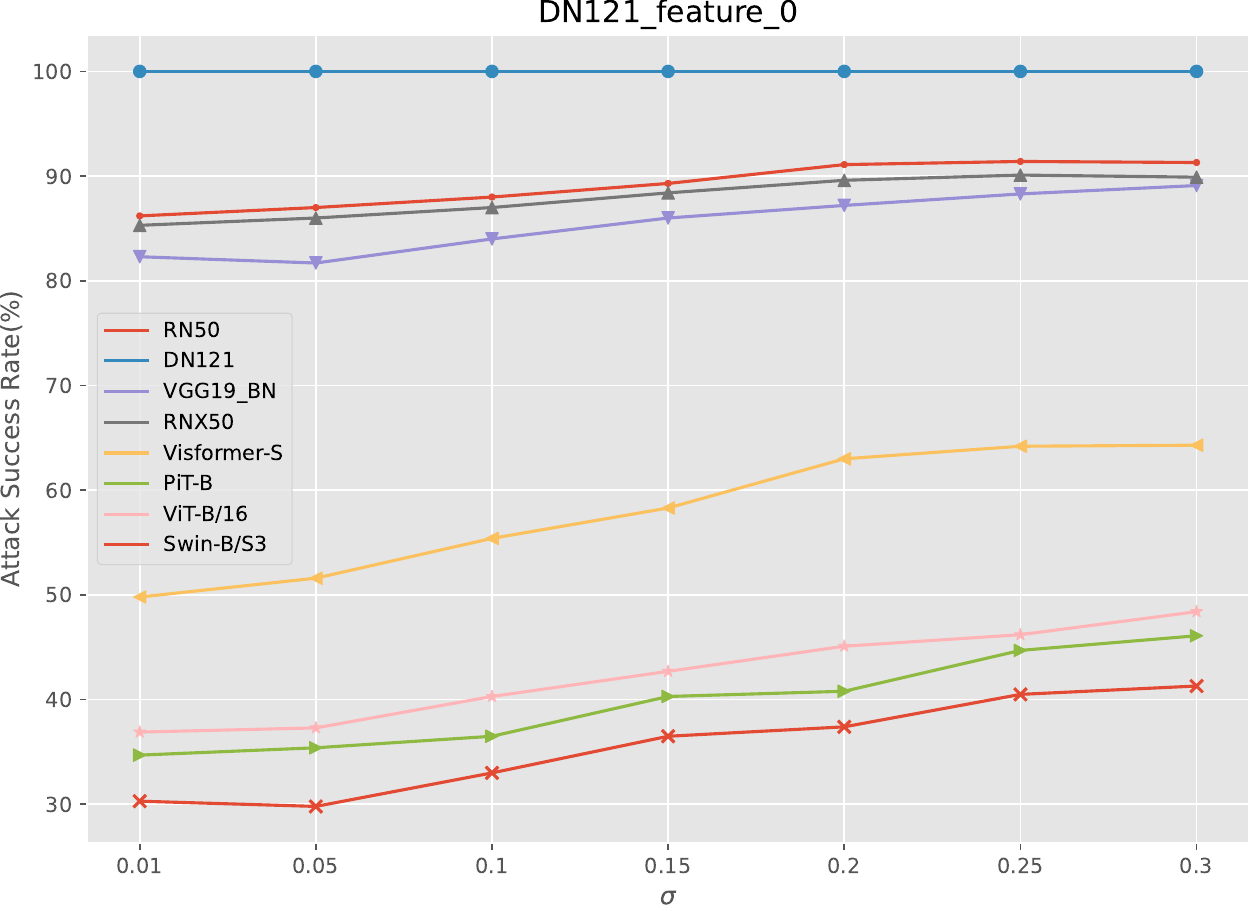}
	\end{minipage}
	\begin{minipage}{.22\linewidth}
		\centering
		\includegraphics[width =1.\linewidth]{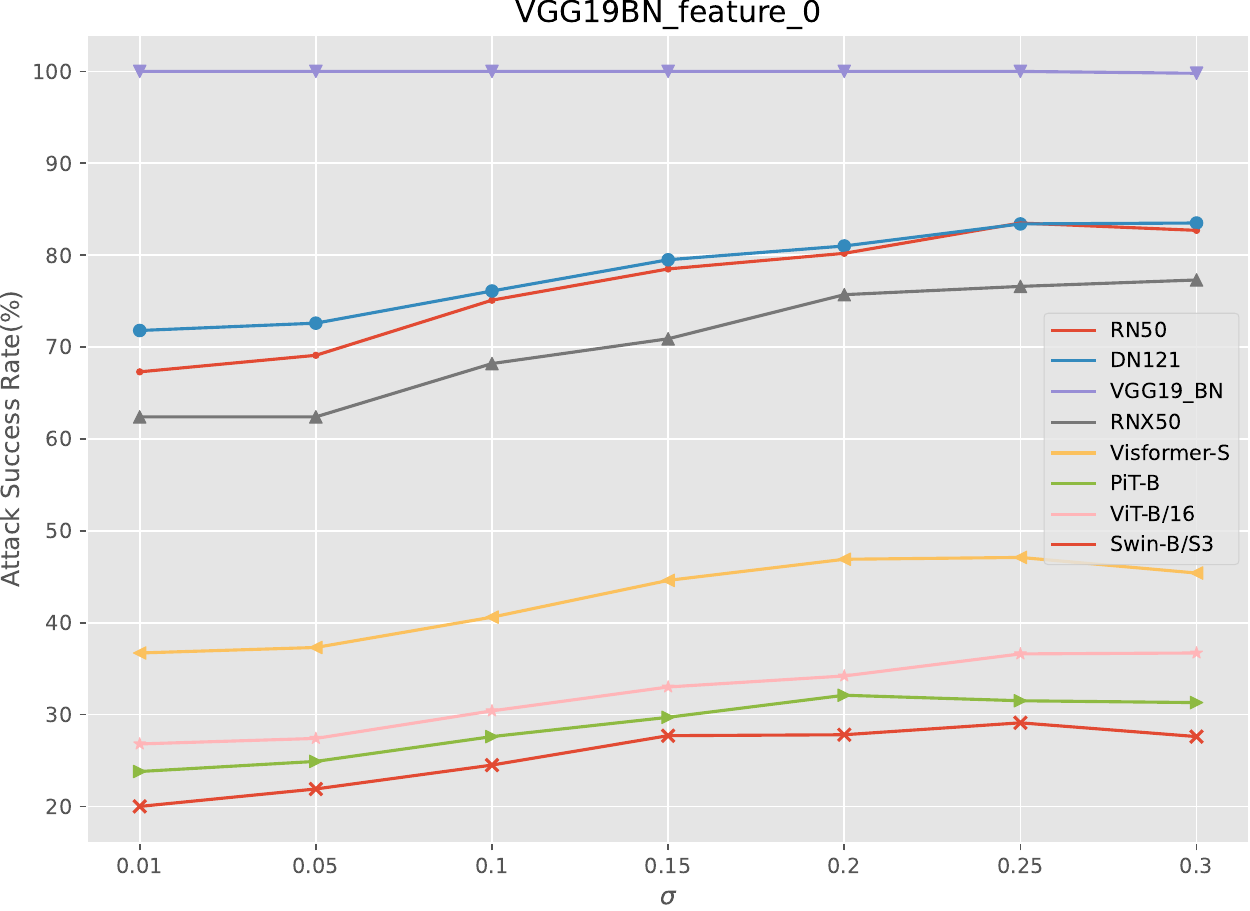}
	\end{minipage}
	\begin{minipage}{.22\linewidth}
		\centering
		\includegraphics[width =1.\linewidth]{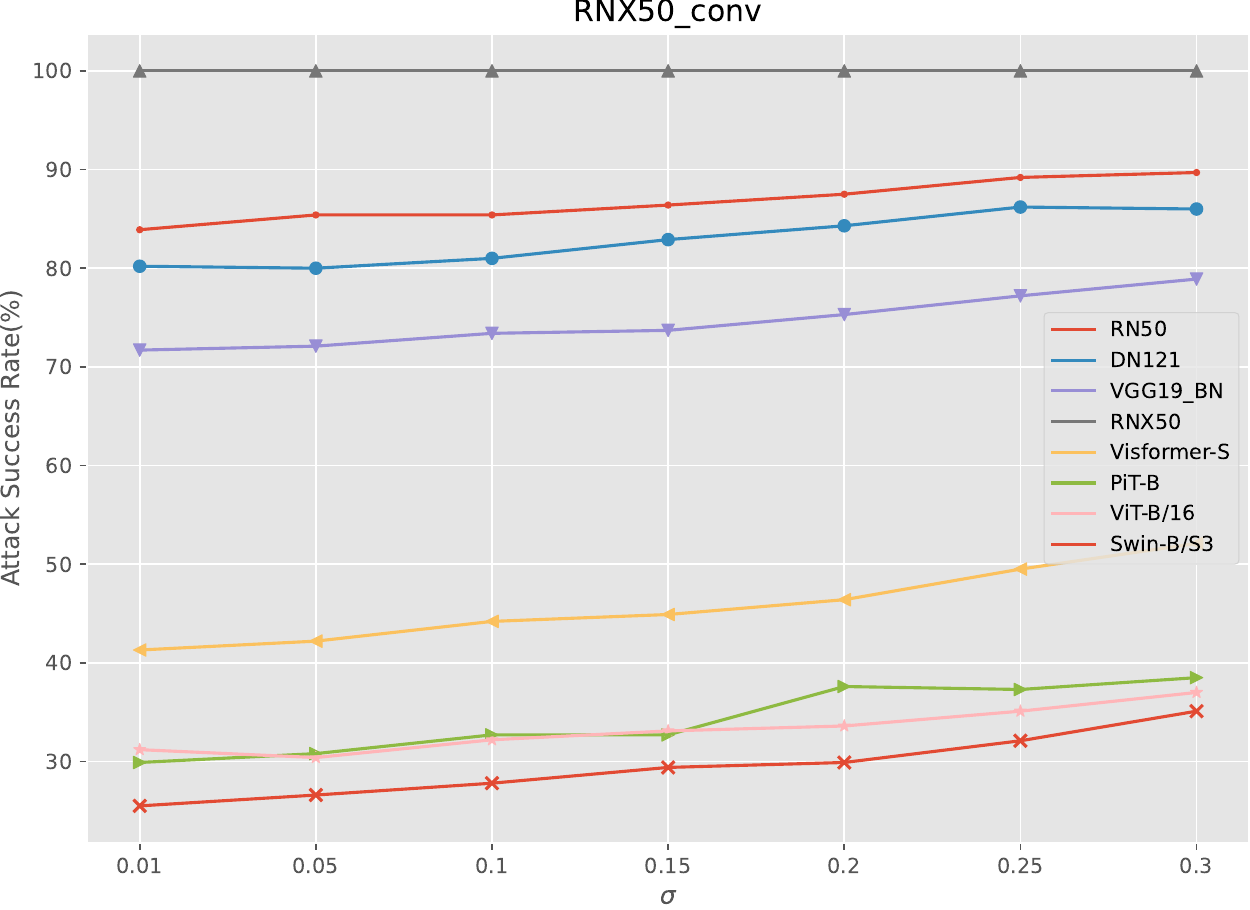}
	\end{minipage}
	
	\begin{minipage}{.22\linewidth}
		\centering
		\includegraphics[width =1.\linewidth]{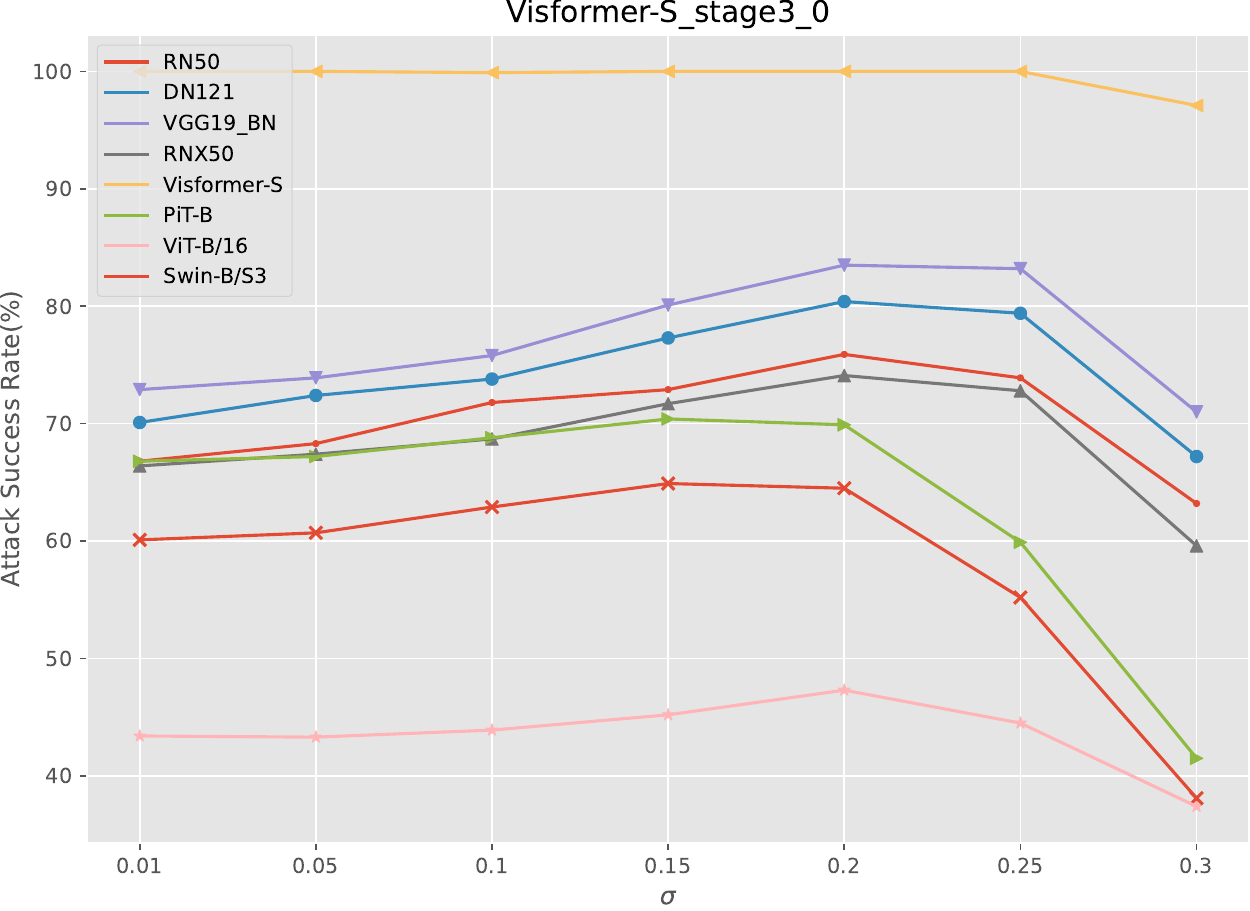}
	\end{minipage}
	\begin{minipage}{.22\linewidth}
		\centering
		\includegraphics[width =1.\linewidth]{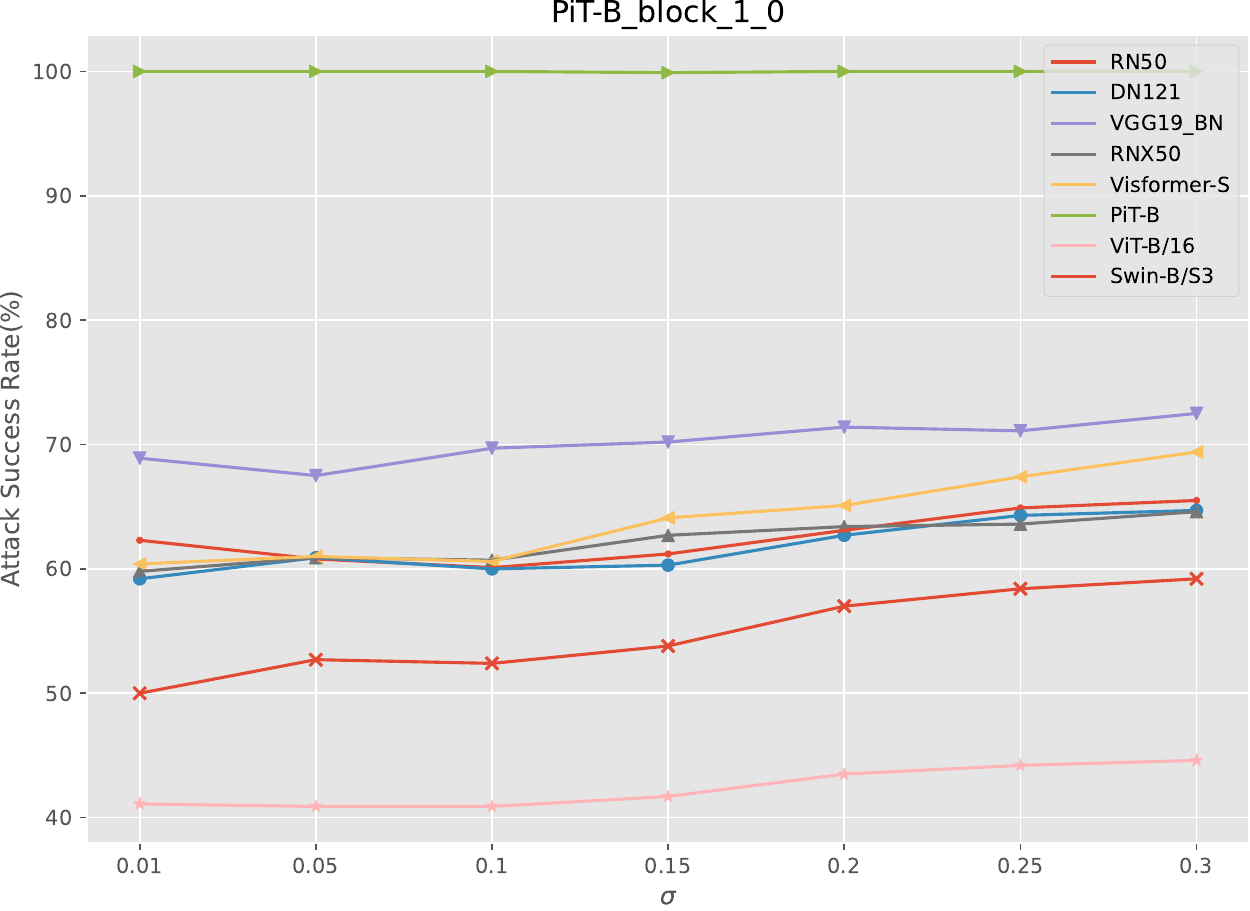}
	\end{minipage}
	\begin{minipage}{.22\linewidth}
		\centering
		\includegraphics[width =1.\linewidth]{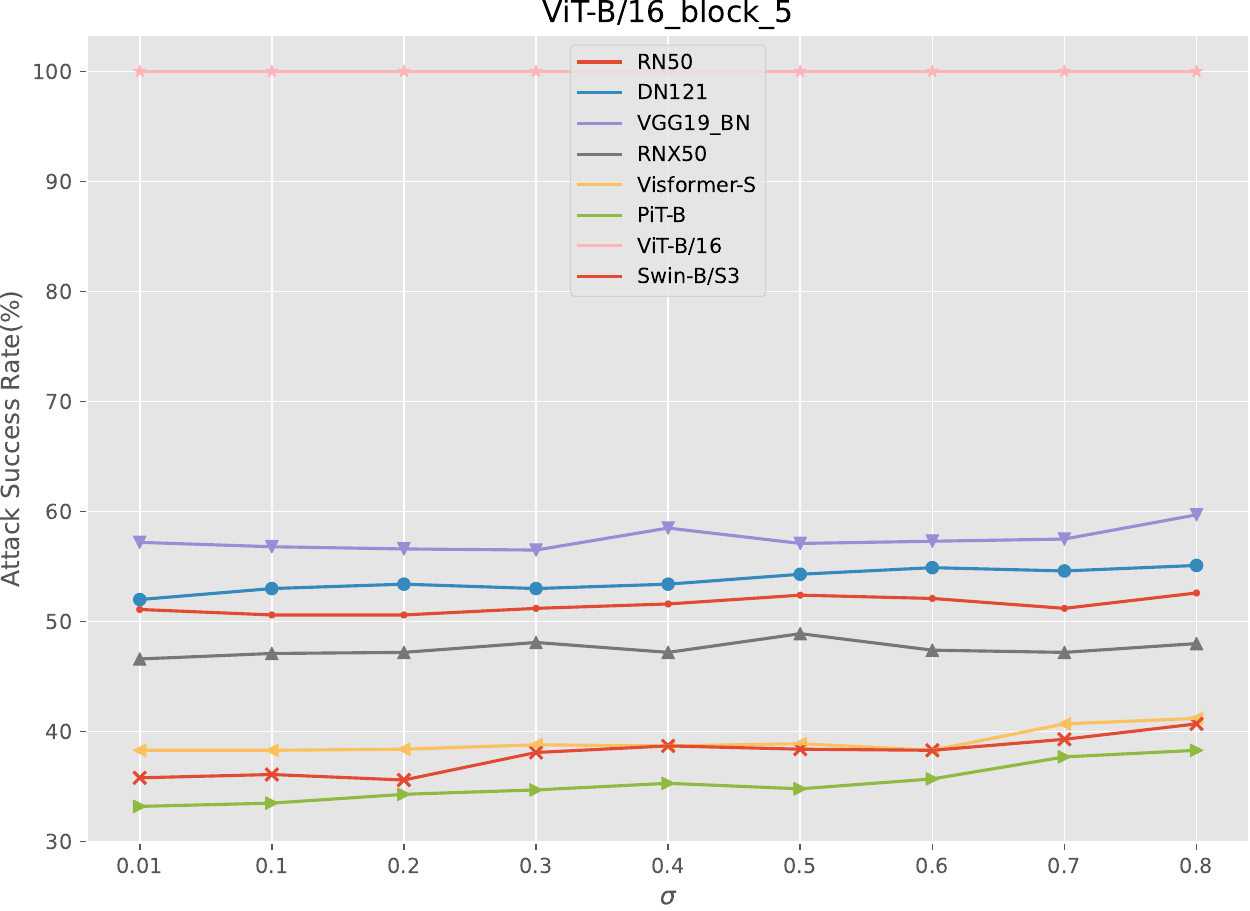}
	\end{minipage}
	\begin{minipage}{.22\linewidth}
		\centering
		\includegraphics[width =1.\linewidth]{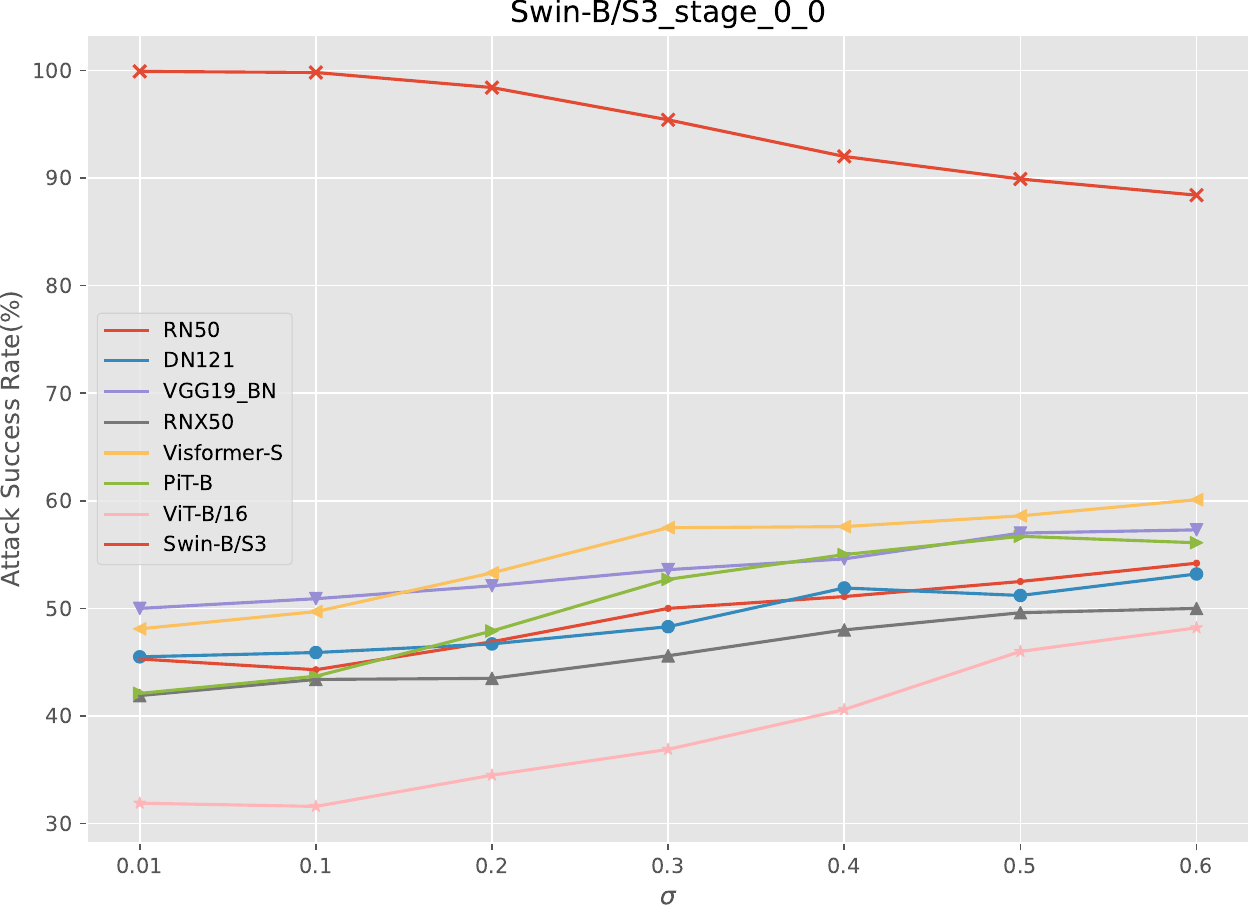}
	\end{minipage}
\caption{Influence of random noise strength on attack performance at the specific layer.}
\label{fig:abla_std}
\end{figure}

In this section, we present an ablation study on the noise strength (i.e., $\sigma$) of feature augmentation across eight models, where the intermediate layer is set with the value reported in Table \ref{supp:tab_param}. The ablation results are illustrated in Figure \ref{fig:abla_std}. At first glance, the adversarial transferability rate increases with the rising noise strength, except for Visformer-S. Meanwhile, some models require a large noise strength to achieve a better adversarial transferability rate; for example, ViT-B/16 and Swin-B/S3 achieve the best average transferability rate of 52.74\% and 54.33\% when $\sigma$ is set to 0.8 and 0.6. The observation suggests that different models exhibit varying degrees of robustness to the random noise under different noise strengths in feature space. The evaluation result indicates that a small noise strength is insufficient to affect the features of some models and has limited influence on avoiding overfitting. In contrast, large noise strength has a greater probability of avoiding overfitting.

%
%